\documentclass{article}
\usepackage[final]{neurips_2021}
\usepackage[utf8]{inputenc} % allow utf-8 input
\usepackage[T1]{fontenc}    % use 8-bit T1 fonts
\usepackage[]{hyperref}       % hyperlinks
\usepackage{booktabs}       % professional-quality tables
\usepackage{amsfonts}       % blackboard math symbols
\usepackage{nicefrac}       % compact symbols for 1/2, etc.
\usepackage{microtype}      % microtypography
\usepackage{nicefrac}
\usepackage{amssymb}
\usepackage{amsmath}
\usepackage{amsthm}
\usepackage{cleveref}
\usepackage{autonum}
\usepackage{MnSymbol} 

\usepackage{caption}
\usepackage{url}
\usepackage{lettrine}
\usepackage{framed}
\usepackage{color}

\usepackage{algorithm,algpseudocode}

\algnewcommand{\algorithmicforeach}{\textbf{for each}}
\algdef{SE}[FOR]{ForEach}{EndForEach}[1]
  {\algorithmicforeach\ #1\ \algorithmicdo}% \ForEach{#1}
  {\algorithmicend\ \algorithmicforeach}% \EndForEach

\usepackage{overpic}
\usepackage{dsfont}
\usepackage{booktabs} 
\usepackage{bm}
\usepackage{bbm}
\usepackage{ifthen}
\usepackage{graphicx}
\usepackage{subcaption}
\usepackage{dsfont}
\usepackage{framed}
\usepackage{overpic}
\usepackage[svgnames]{xcolor}
\newtheorem{theorem}{Theorem}
\usepackage{apxproof}
\newtheorem{proposition}{Proposition}
\newtheorem{definition}{Definition}
\newtheorem{lemma}{Lemma}
\newtheorem{assumption}{Assumption}
\newtheorem{corollary}{Corollary}[theorem]
\usepackage{xr}
\hypersetup{
	colorlinks=true,
	linkcolor=blue,
	urlcolor=red,
	citecolor=DarkBlue,
	linkbordercolor={0 0 1}
}

\newtheoremrep{theorem}{Theorem}
\renewcommand{\algorithmicrequire}{\textbf{Input:}}
\renewcommand{\algorithmicensure}{\textbf{Output:}}

\newcommand{\edit}[1]{{\color{red} (AS: #1)}}

\theoremstyle{definition}
\newtheorem{example}{Example}[section]
\title{No-regret Algorithms for Fair Resource Allocation}
\author{
  Abhishek Sinha, Ativ Joshi \\
 School of Technology and Computer Science \\
  Tata Institute of Fundamental Research \\
  Mumbai 400005, India \\
  \texttt{abhishek.sinha@tifr.res.in} \\
  \texttt{ativ@cmi.ac.in}
   \And
   Rajarshi Bhattacharjee, Cameron Musco,\\
   \textbf{Mohammad Hajiesmaili}\\
   Manning College of Information \\and Computer Sciences\\
  University of Massachusetts Amherst\\
  \texttt{\{rbhattacharj, cmusco,}\\\texttt{hajiesmaili\}@cs.umass.edu}
}

\begin{document}

\maketitle

\begin{abstract}
We consider a fair resource allocation problem in the no-regret setting against an unrestricted adversary. The objective  is to allocate resources equitably among several agents in an online fashion so that the difference of the aggregate $\alpha$-fair utilities of the agents between an optimal static clairvoyant allocation and that of the online policy grows sub-linearly with time. The problem is challenging due to the non-additive nature of the $\alpha$-fairness function. Previously, it was shown that no online policy can  exist for this problem with a sublinear standard regret. In this paper, we propose an efficient online resource allocation policy, called Online Proportional Fair (\texttt{OPF}), that achieves $c_\alpha$-approximate sublinear regret with the approximation factor $c_\alpha=(1-\alpha)^{-(1-\alpha)}\leq 1.445,$ for $0\leq \alpha < 1$. The upper bound to the $c_\alpha$-regret for this problem exhibits a surprising \emph{phase transition} phenomenon. The regret bound changes from a power-law to a constant at the critical exponent $\alpha=\nicefrac{1}{2}.$  As a corollary, our result also resolves an open problem raised by \citet{even2009online} on designing an efficient no-regret policy for the online job scheduling problem in certain parameter regimes. The proof of our results introduces new algorithmic and analytical techniques, including greedy estimation of the future gradients for non-additive global reward functions and bootstrapping adaptive regret bounds, which may be of independent interest.
\end{abstract}

%\keywords{Online learning, fairness }
\section{Introduction} \label{intro}
%\Cam{I feel that maybe we should focus less on fair learning and just focus on fair allocation from the beginning? E.g. we could just start with the caching example.}
%\edit{That's surely a possibility. I wanted to write a broad introduction suitable for the general NeurIPS audience and place the fair resource allocation problem in a broader context. I felt talking about the caching problem right at the introduction would narrow the paper's scope. The ad example given below is a specific case of the scheduling problem which we do study in this paper.}
The notion of \emph{algorithmic fairness} refers to  learning algorithms that guarantee fair predictions even when subjected to adversarially-biased training data \citep{dwork-fairness}. In addition to efficiency, fairness has become a major criterion for designing and deploying large-scale learning algorithms that affect a diverse user base. Since the training data could be highly skewed in practice, it is essential to make minimal assumptions about the data-generating process and design provably robust fair learning policies. Guaranteeing fairness becomes even more challenging in the online learning set-up as there is no distinction between the training and test data, and no assumption is made on the input data sequence. As an example, consider an online recruitment campaign where a learning algorithm decides the target group (identified by, say, the tuple (\texttt{race}, \texttt{gender}, \texttt{age})) to which an ad for a job vacancy is to be displayed. Suppose a revenue-maximizing recommendation algorithm concludes from past data that more revenue is generated by showing the ad to Group A compared to Group B. In that case, it will eventually end up showing that ad exclusively to Group A while discriminating against Group B users of a potential job opportunity \citep{hao2019facebook}. One of the overarching goals of this paper is to design efficient online learning policies that provably mitigate this algorithmic bias while maintaining  efficiency \emph{irrespective} of the past data seen so far (refer to Example \ref{sched-example} for a relevant model for the above example considered in this paper).  
 %Specifically, while matching jobs to the users in a crowdsourcing scenario, it is important  
 
Towards this goal, we consider a generic online fair resource allocation problem called \texttt{NOFRA} (\underline{No}-Regret \underline{F}air \underline{R}esource \underline{A}llocation). In this problem, a fixed set of resources needs to be equitably shared among $m$ agents over multiple rounds. Note that fairness is a complex, multidimensional, and essentially subjective concept. Several quantitative metrics have been introduced in the literature for quantifying the degree of fairness in resource allocation, including $\alpha$-fairness \citep{lan2010axiomatic}, proportional fairness \citep{kelly1997charging, mo2000fair}, max-min fairness \citep{radunovic2007unified, nace2008max}, and Jain's fairness index \citep{jain1984quantitative}. In this paper, we consider the problem of maximizing the $\alpha$-fairness function, in which the utility of each agent when allocated $R$ units grows as $R^{1-\alpha}$. The parameter $\alpha$ is restricted to the interval $[0, 1).$ This same range of $\alpha$ has been studied earlier in a game-theoretic set up by \citet{ALTMAN2010338}.
 
 \paragraph{On the hardness of the \texttt{NOFRA} problem:} \label{difficulty}

%We now outline the fundamental difference between the \texttt{NOFRA} problem \eqref{regret-def} and the standard online learning framework . 
In the standard online learning setting, the cumulative reward accrued over a given time horizon is taken to be the sum of the rewards obtained at each round \citep{hazan2019introduction}. In contrast, the objective of the \texttt{NOFRA} problem is to maximize the global $\alpha$-fairness function, which is equal to the sum of cumulative rewards of each agent raised to the power $1-\alpha, 0\leq \alpha <1$. Because of the power-law non-linearity, the objective function of the \texttt{NOFRA} problem is non-additive with respect to time. Note that the non-additivity of the objective function is essential to induce fairness in sequential allocations by incorporating the diminishing return property. However, this renders the \texttt{NOFRA} problem fundamentally different from standard online learning problems. In particular, Theorem \ref{lb-thm} proves a non-trivial lower bound to the approximation ratio achievable by any online learning policy for this problem. This lower bound does not hold in classic settings with additive rewards.
%shows that it is impossible to design an no-regret algorithm approximation factor better than $1.06$. 
Although some specific problems with non-additive reward functions have been previously studied in the online learning literature \citep{si2022enabling, even2009online, rakhlin2011online}, in the following, we explain why the \texttt{NOFRA} problem is fundamentally different from the existing studies.

%\edit{Talk about algorithmic fairness and explain how the notion of fairness makes a difference in the perceived utility of the users. Take the example of shared caching problem with two users and consider the case when one user requests the same file and the other user requests a wide variety of files.}.  
\paragraph{Related work:} 
In a closely-related paper, \citet{si2022enabling} considered the problem of designing fair online resource allocation policies. The authors showed that it is impossible to design a no-regret policy without restricting the set of admissible adversarial demand sequences. Given this negative result, the authors proposed a no-regret policy using a primal-dual framework under the assumption of a restricted adversary, which exhibits an essentially i.i.d. random-like $o(T)$ fluctuation. In contrast, we design a robust policy with an approximate sublinear regret \emph{without restricting} the adversary. Due to the weaker assumption, our policy and its analysis are also very different from that of \citet{si2022enabling}. \citet{online-fair} considered an online resource allocation problem where the objective is to guarantee a sublinear regret for the allocation efficiency and a sublinear minimum guarantee violation penalty. The authors proposed an online policy that achieves this goal by using a weighted $\alpha$-fair allocation on each round while sequentially tuning the weights and the exponents of the $\alpha$-fairness function. Although they use round-wise $\alpha$-fair allocation as a tool, their objective is not to optimize the $\alpha$-fairness of the cumulative allocation, which is the focus of our paper. \cite{even2009online} considered optimizing a global concave objective function in the no-regret setup, under the further assumption that the optimal reward is convex. They presented an approachability-based policy in this setting and also proved the impossibility of achieving zero regret when the convexity condition is violated. In our case of the $\alpha$-fairness function for $0\leq \alpha < 1$, although the objective function is concave, we show that the optimal static offline reward function fails to be convex (see Appendix \ref{non-cvx}). Hence, their policy does not apply to the \texttt{NOFRA} problem. The paper by \citet{rakhlin2011online} considered the problem of no-regret learnability for a wide class of non-additive global functions. However, their results also do not apply to our setting as our problem is not no-regret learnable (see Theorem \ref{lb-thm}). 
Several prior works exist that design no-regret policies for specific non-additive functions. As an example, 
\citet{blum1997line, fine-paging} used online learning techniques for solving the online paging and the Metrical Task System problem, which contains states. The fairness problem has also been extensively studied in the stochastic multi-armed bandit setting \citep{narahari-fair, joseph2016fairness, li2019combinatorial}. 

%considered the \texttt{Fair-MAB} problem where the objective is to minimize the regret for the sum of expected rewards under the constraint that each arm is pulled a pre-specified number of times. 
 %In particular, the reward accrued at any round $t$ depends on the action taken on round $t$ \emph{only}. 
%Due to the separable rewards structure, the standard online learning framework applies to \emph{memoryless} settings where the action taken at a round does not affect the reward functions seen in the subsequent rounds. On the other hand, i
%In the \texttt{NOFRA} problem, the action $\bm{y}_t \in \Delta$ taken on round $t$ affects the final rewards through the cumulative reward vector $\bm{R}(t)$. For the same reason, recent online algorithms developed for systems with \emph{bounded memory} \citep{anava2015online} do not apply to the \texttt{NOFRA} problem. 
%From a high-level perspective, the online fair caching problem is an instance of an adversarial Reinforcement Learning problem, for which no general algorithmic methodology is known. 
%\subsection{Our contributions} \label{contributions}

%\Cam{Doesn't need to be done for arxiv but it would be nice to convey here that our algorithm is simple an intuitive: we have weights on each user that decay as the user's cumulative reward increases.}
\paragraph{Our contributions:}
% In view of the hardness of the problem and prior works, 
 We show that despite its non-additive structure, the \texttt{NOFRA} problem can be approximately reduced to an instance of an online linear optimization problem with a greedily defined sequence of reward vectors. In particular, we make the following contributions:
\begin{itemize}
	\item In Algorithm \ref{OPFC-main}, we present an efficient online resource allocation policy, called Online Proportional Fair (\texttt{OPF}), that approximately maximizes the aggregate $\alpha$-fairness function of the agents in the sense of standard regret. We show that the above policy achieves $(1-\alpha)^{-(1-\alpha)} \leq 1.445$-approximate sublinear regret (Theorem \ref{main-result}). To the best of our knowledge, \texttt{OPF} is the first online policy that approximately maximizes the $\alpha$-fairness for any adversarial sequence.
	\item  In Theorem \ref{lb-thm}, we establish a lower bound to the approximation factor $c_\alpha$ achievable by any online learning policy for the $\alpha$-fair reward function. Our lower bound improves upon the best-known lower bound for this problem.
 %In Theorem \ref{lb-thm}, we show that the achievable approximation factor $c$ for any online policy with a sublinear $c$-regret is lower bounded by $1.06$ for $\alpha=\nicefrac{1}{2}.$ 
	\item On the algorithmic side, we introduce a new class of online policies for optimizing non-additive global reward functions by greedily estimating the future gradients and using these estimated gradients within an online gradient ascent policy with adaptive step sizes. The resulting algorithm is simple and intuitive: we have weights on each user that decay as the user's cumulative reward increases. We show that the greedy estimation is sufficient for obtaining a constant-factor approximate sublinear regret for the $\alpha$-fair reward function.
	\item On the analytical side, we introduce a new proof technique that simultaneously controls the magnitude of the gradients and the adaptive regret bound under the above policy. This technique applies to problems with memory or states where the future gradients depend on past actions. 
	%\item We demonstrate the efficacy of the proposed policy through extensive numerical experiments with synthetic and real-world datasets. 
\end{itemize}

\section{Problem Formulation} \label{gen}
%We consider the \underline{O}\underline{n}line \underline{F}air \underline{R}esource \underline{A}llocation (\texttt{NOFRA}) problem in a no-regret set up.
In this section, we give a general formulation of the \texttt{NOFRA} problem. In Section \ref{examples}, we give examples of  three concrete resource allocation problems that fit into this general framework.

Let the term \emph{agents} denote entities among which a limited resource is to be fairly divided. Assume that the resource allocated to the $i$\textsuperscript{th} agent on the $t$\textsuperscript{th} round is represented by an $N$-dimensional non-negative Euclidean vector $y_i(t)$, where $N$ is an arbitrary number.
On every round $t$, the $i$\textsuperscript{th} agent requests an $N$-dimensional non-negative \emph{demand} (or, reward) vector $x_i(t)$. The demand vectors are revealed to an online allocation policy $\pi$ at the end of each round. We make \emph{no assumption} on the regularity of the demand vector sequence, which could be adversarially chosen (\emph{c.f.,} \cite{si2022enabling}). Before the $N \times m$ dimensional aggregate demand matrix $\bm{x}(t) \equiv \big(x_1(t), x_2(t), \ldots, x_m(t)\big)$ for round $t$ is revealed, the online resource allocation policy $\pi$ chooses a non-negative $N \times m$ dimensional allocation matrix $\bm{y}(t)=\big(y_1(t), y_2(t), \ldots, y_m(t)\big)$ from the set of all feasible allocations $\Delta$. The set of all feasible allocations is assumed to be convex (see Remark 2 below).
%The set of all allocations $\Delta$ is assumed to be convex. 
%The component vectors $x_i(t)$ and $y_i(t)$ have the same dimension $N$ (say), and are element wise non-negative. 
The reward accrued by the agent $i$ on round $t$ is given by the inner-product $\langle x_i(t), y_i(t)\rangle,$ which, without any loss of generality, is assumed to be upper-bounded by one.
% which, in reference to the caching problem described later, is referred to as a \emph{Hit}. 
The total cumulative reward accrued by agent $i$ agent at the end of round $t$ is given by 
\begin{eqnarray} \label{recursion}
	R_i(t+1) = R_i(t) + \langle x_i(t), y_i(t)\rangle, ~ \forall i, t,  
\end{eqnarray}
where $R_i(0)=0, \forall i. $ By iterating the above recursion, the cumulative reward $R_i(t)$ can be alternatively expressed as follows:
\begin{eqnarray} \label{hit-eqn}
	R_i(t) = \sum_{\tau=1}^{t-1} \langle x_i(\tau), y_i(\tau) \rangle.
\end{eqnarray}
We make two mild technical assumptions on the structure of the demand and allocation vectors.
\begin{framed}
\begin{assumption}\label{assumption1}
	$\delta \leq ||x_i(t)||_1 \leq 1, \forall i,t,$ for some constant $\delta>0.$
\end{assumption}
%In other words, the reward vectors are strictly positive.

\begin{assumption}\label{assumption2}
Let $\bm{1}_{N\times m}$ denote the $N\times m$ all-$1$ matrix. Then $\mu \bm{1}_{N\times m} \in \Delta$ for some constant $\mu>0.$  
\end{assumption}
\end{framed}
The above assumptions imply that it is possible to ensure a non-zero reward for all agents on all rounds. This property will be exploited in the proofs of our regret bounds (see Eq.\ \eqref{oco-prop}). 
%From Eqn.\ \eqref{hit-eqn1}, it is clear that cumulative reward $R_i(t)$ accrued by the $i$\textsuperscript{th} agent up to time $t,$ \emph{i.e.,} is a function of only the first $t$ action variables $\{y_\tau\}_{\tau=1}^t.$ 

The utility of any user for a cumulative reward of $R$ is given by the concave $\alpha$-fair utility function $\phi: \mathbb{R}_+ \to \mathbb{R}_+,$ defined as follows: 
%In this paper, we will exclusively work with the $\alpha$-fair utility function 
\begin{eqnarray} \label{alpha-fair}
	\phi(R)=\phi_\alpha(R)\equiv  \frac{R^{1-\alpha}}{1-\alpha}, ~~R\geq 0,
\end{eqnarray}
for some constant $0\leq \alpha <1$.\footnote{Some authors define the $\alpha$-fairness function with an extra additive term, \emph{i.e.,} $\phi'_\alpha(R) = \frac{R^{1-\alpha}-1}{1-\alpha}, \alpha \neq 1$ and $\phi'_\alpha(R) =\ln(R), \alpha=1$ \citep{si2022enabling}. Note that, in the range $0\leq \alpha <1$, this alternative definition changes the regret metric \eqref{regret-def} only by an additive constant. We use the fairness function $\phi_\alpha(R)$ as it is \emph{positively homogeneous} of degree $(1-\alpha)$ - a property that we exploit in our analysis.} The fairness parameter $\alpha$ induces a trade-off between the desired efficiency and fairness by incorporating a notion of \emph{diminishing return} property in the global objective function. The static offline optimal allocation with larger $\alpha$ leads to more equitable cumulative rewards \citep{bertsimas2012efficiency}. Setting $\alpha=0$ reduces the problem to the ``unfair'' online linear optimization problem. 
%\textcolor{red}{Explain the rationale for choosing $\alpha$-fair utility and add references.}
Our objective is to design an online resource allocation policy $\pi \equiv \{\bm{y}(t)\}_{t \geq 1}$ that minimizes the regret for maximizing the aggregate utilities of all users compared to any fixed offline resource allocation strategy $\bm{y}^* \in \Delta.$
%which could be a convex combination of integral allocations
 To be precise, assume that the offline, fixed resource allocation $\bm{y}^*$ yields a cumulative reward  of $\bm{R}^*(T)$ at the end of round $T\geq 1.$ Then, our objective is to design a resource allocation policy which minimizes the $c$-approximate regret (which we refer to as $c$-regret for short) defined as:
\begin{eqnarray} \label{regret-def}
	\textrm{Regret}_T(c) \equiv \sum_{i=1}^m \phi(R_i^*(T))-   c\sum_{i=1}^m \phi(R_i(T)),
\end{eqnarray}
for some small constant $c\geq 1.$\footnote{Our theoretical results and algorithms trivially extend to the $\bm{w}$-weighted fairness function $\sum_{i}w_i \phi(R_i(T))$ for some non-negative weight vector $w_i \geq 0, \forall i.$} In the case of standard regret ($c=1$), we drop the argument in the parenthesis.
%We explore the dynamic allocation of the proposed \texttt{OPF} policy in the regime $\alpha>1$ through numerical experiments. 
Note that, unlike the standard online convex optimization problem, in the \texttt{NOFRA} problem, the reward function is global, in the sense that it is non-separable across time \citep{even2009online}. For this reason, it is necessary to consider the $c$-regret with $c > 1$, rather than the standard regret (i.e., with $c = 1$). This will become clear from Theorem \ref{lb-thm}, where we prove an explicit lower bound to the approximation factor achievable by any online policy for the global $\alpha$-fair reward function. This lower bound implies a concrete lower bound on the achievable $c$.
%lower bound the achievable approximation factor for all online policies.
%we show that for all online policies with a sublinear $c_\alpha$-regret for the $\alpha=\nicefrac{1}{2}$-fair utility function, the approximation factor must be at least $1.06.$ 
Our lower bound improves upon   \cite[Theorem 1]{si2022enabling}, where it was shown that no online policy can achieve a sublinear standard regret for the \texttt{NOFRA} problem under an unrestricted adversary.

%We emphasize that the horizon length $T$ is not necessarily known to the online policy a priori. Hence, we seek an online policy with an \emph{anytime} regret guarantee that achieves near-optimal fairness \eqref{regret-def} on \emph{all} rounds $T \geq 1$. 

\smallskip

\noindent \textbf{Remark 1:} When $\alpha>1,$ the offline benchmark $\phi(R_i^*(T))$ itself becomes $O(1), \forall i.$ Hence, in this regime, a sublinear regret bound \eqref{regret-def} becomes vacuous. Consequently, we restrict the fairness parameter $\alpha$ to the interval $[0, 1).$ 

\smallskip

\noindent \textbf{Remark 2:} Apart from Section \ref{int-alloc}, we assume that the set of feasible allocations $\Delta$ is convex and thus that there are no integrality constraints on the allocation, \emph{i.e.,} the components of the allocation matrix $\bm{y}(t)$ are allowed to be fractional. However, in many combinatorial resource allocation problems, such as shared caching (Example \ref{caching-examp}), the allocation vector is required to be integral. In this case, the feasible action set $\Delta$ is naturally defined to be the convex hull of the integral actions. In Section \ref{int-alloc}, we consider the integrality constraints on the allocation vector and derive a randomized integral allocation policy with a sublinear regret bound as a corollary of our results for the relaxed problem.

\subsection{Examples}\label{examples}

The statement of \texttt{NOFRA} is fairly general and by suitably choosing the reward and allocation vectors, many resource allocation problems can be reduced to \texttt{NOFRA}. In this section, we highlight three such problems: fair allocation in online shared caching, online job scheduling, and online matching. %, can be formulated within the \texttt{NOFRA} framework.

%\paragraph{Example 1: Online Caching (Single Cache) \citep{SIGMETRICS20}:} 
\begin{example}[Online Shared Caching \citep{SIGMETRICS20}]\label{caching-examp}
%We now show that the \emph{Online Fair Caching} problem can be considered as a special case of the general \texttt{NOFRA} problem. 

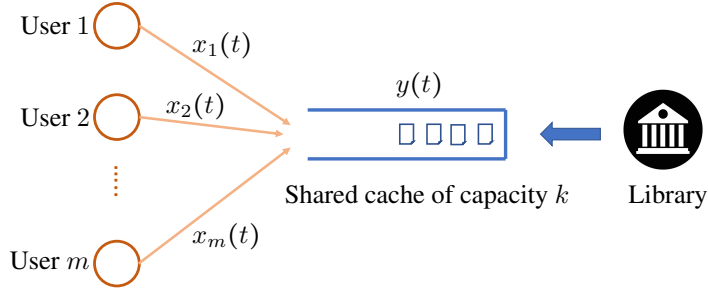
\begin{figure}[!ht]
\centering
	\includegraphics[scale=0.3]{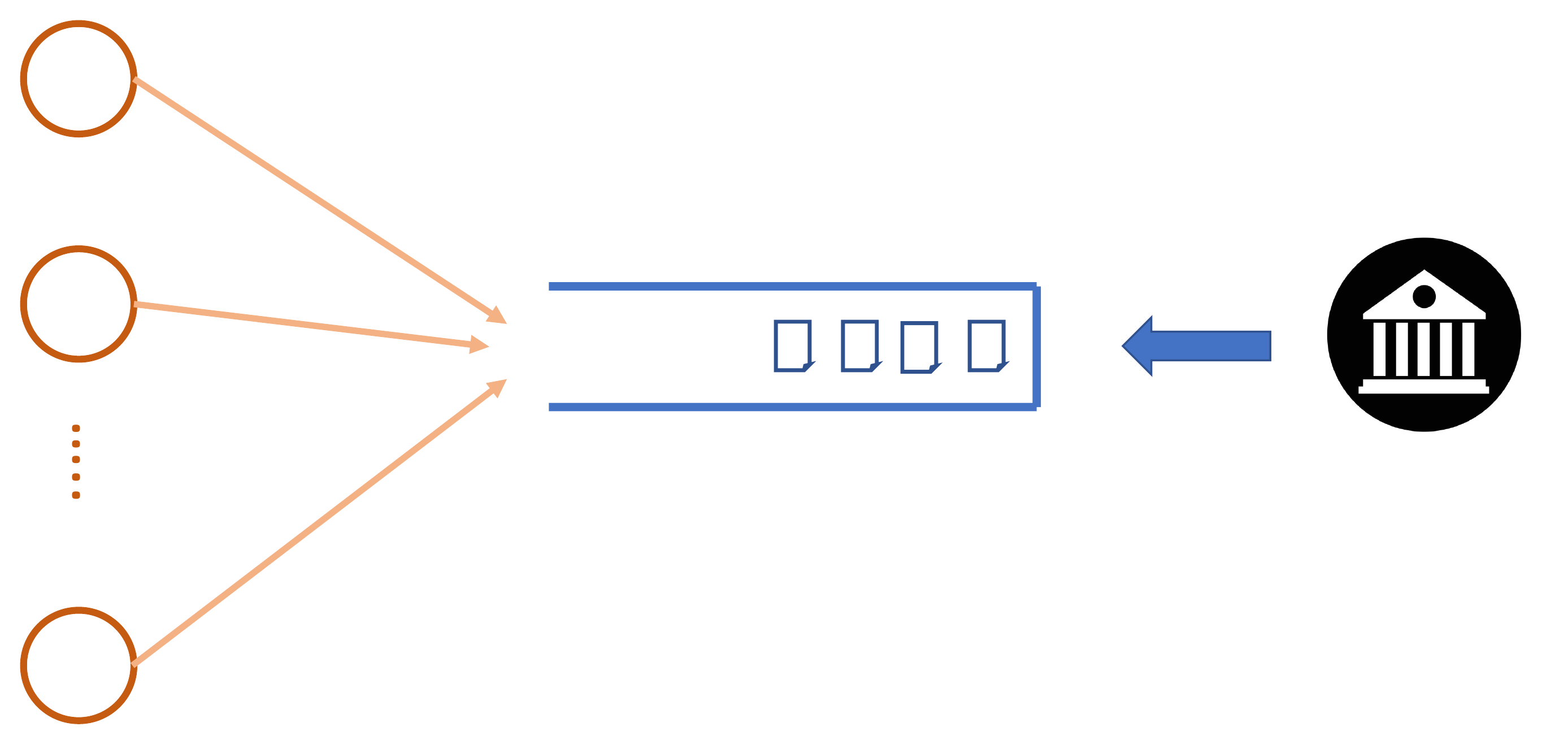}
	\put(-35, 35){Library}
	\put(-200, 92){$\tiny{x_1(t)}$} 
	\put(-210, 69){$\tiny{x_2(t)}$} 
	\put(-200, 20){$\tiny{x_m(t)}$} 
	\put(-123, 77){$\tiny{y(t)}$}
	\put(-165,35){Shared cache of capacity $k$}
	\put(-265,99){User $1$}
    \put(-265,64){User $2$}
    \put(-269,10){User $m$}
	\caption{Online shared caching problem} 
	\label{sys-diagram}
\end{figure}

In the \emph{Online Shared Caching} problem with a single shared cache, $m$ users are connected to a single cache of capacity $k$ (see Figure~\ref{sys-diagram} for a schematic). At each round, each user requests a file from a library of size $N.$ The file request sequence may be adversarial. At the beginning of each round $t$, an online caching policy prefetches at most $k$ files on the cache, denoted by the vector $y(t)$ such that 
\begin{eqnarray} \label{caching-feasible}
	\sum_{i=1}^N y_i(t)=k, ~~ 0\leq y_i(t) \leq 1~, \forall i \in [m].
\end{eqnarray}
 The set of all feasible caching configurations is denoted by $\Delta^N_k.$\footnote{In the above, we allow fractional caching, which can be easily converted to a randomized integral caching policy (where $y_i(t)\in \{0,1\}, \forall i,t$) via sampling. See Section \ref{int-alloc} for details.}
%\footnote{In the following, we consider the relaxed action where the components of any vector $y \in \Delta^N_k$ denote the corresponding inclusion probabilities. From the any relaxed action, one can derive an integral caching policy by using Madow's sampling with these probabilities. As a result, the variable $R_i(t)$'s in the regret definition are to be taken as the expected hit rates (\emph{not} the actual hits), which can be computed by the policy from the inclusion probabilities $y(t)$ as in Eqn.\ \eqref{reward-evolution}.}.  
Immediately after the prefetching at round $t$, each user $i$ reveals its file request, represented by the one-hot encoded demand vector $x_i(t)$, $1\leq i \leq m$. A special case of the above caching model for a single user ($m=1$) has been investigated in previous work \citep{SIGMETRICS20, mhaisen2022optimistic, ITW22}. 

Let $R_i(t)$ denote the cumulative (fractional) hits obtained by the $i$\textsuperscript{th} user up to round $t \geq 1$. Clearly,
\begin{eqnarray} \label{reward-evolution}
	R_i(T) = \sum_{t=1}^T\langle x_i(t), y(t) \rangle, ~~1 \leq i \leq m,
\end{eqnarray}
%where $R_i(0)=0, \forall i. $ 
%The objective of the Online Caching problem is to design an online caching policy that minimizes the $c$-Regret \eqref{regret-def}.
%\paragraph{Reducing \texttt{OFC}$\implies$ \texttt{NOFRA}:}

The online shared caching problem can be easily reduced to an instance of the \texttt{NOFRA} problem by taking the demand matrix to be $\bm{x}(t)=\big(x_1(t), x_2(t), \ldots, x_m(t)\big).$ Since the allocation vector $y(t)$ is common to all users,  the allocation matrix can be taken to be $\bm{y}(t)=\big(y(t), y(t), \ldots, y(t)\big).$ It can be observed that Assumption 2 holds in this case with $\mu=k/N$ by noting that $\frac{k}{N}\bm{1}_N \in \Delta^N_k.$
%\Cam{Should we say why assumption 2 holds?}

\end{example}

\begin{example}[Online Job Scheduling \citep{even2009online}] \label{sched-example}
In this problem, there are $m$ machines, which play the role of agents. A single job arrives at each round. The reward accrued by assigning the incoming job at round $t$ to the $i$\textsuperscript{th} machine is given by $x_i(t)$ where $x_i(t) \in [\delta, 1], \forall i\in [m],$ where  $\delta >0$ is a small positive constant. Before the rewards for round $t$ are revealed, an online allocation policy selects a probability distribution $\bm{y}_t$ on $m$ machines such that \[ \sum_{i=1}^m y_i(t)=1,~ y_i(t)\geq 0, \forall i \in [m].\] The policy allocates a fraction $y_i(t)$ of the job to the $i$\textsuperscript{th} machine $\forall i\in [m]$. As a result, the $i$\textsuperscript{th} machine accrues a reward of $x_i(t)y_i(t)$ on round $t.$ Hence, the cumulative reward accrued by the $i$\textsuperscript{th} machine in a time-horizon of length $T$ is given by:
\[R_i(T)= \sum_{t=1}^T x_i(t)y_i(t), ~\forall i \in [m].\]
The objective of the online fair job scheduling problem is to design an online allocation policy that achieves a sublinear regret with respect to the $\alpha$-fairness of the cumulative rewards defined in \eqref{regret-def}. From the above description, it is immediately clear that this problem is a special case of the general \texttt{NOFRA} problem. The Online job scheduling problem occurs in many practical contexts, \emph{e.g.,} in the targeted ad-campaigning problem discussed in the introduction, the ads can be modelled as jobs and different groups of users can be modelled as machines. 
%\edit{What does our bounds imply about the results in \citep{even2009online}?}

\end{example}

 \iffalse
\begin{example}[Online Bipartite Caching \citep{paria2021}]
The above problem can be immediately extended to the multiple cache set up \citep{paria2021}. In this problem, $m$ users are connected to $n$ different caches, each of capacity $k$. The interconnection among the users $(\mathcal{U})$ and the caches $\mathcal{C}$ are given by a bipartite graph $G(\mathcal{U} \cup \mathcal{C}, E).$ Define the virtual action \citet{paria2021}.
\end{example}
\fi
%\paragraph{Example 3: Online Matching:}
\begin{example}[Online Matching]
Consider an $m\times m$ bipartite graph where the vertices on the left denote the agents and the vertices on the right denote the resources. On every round, each agent can be matched with one resource only, where we allow fractional matchings. The $m\times 1$ demand vector of the agent $i$ on round $t$ is denoted by $x_i(t).$ The $j$\textsuperscript{th} component of the demand vector $x_i(t)$ denotes the potential reward accrued by the $i$\textsuperscript{th} agent had it been completely matched with the $j$\textsuperscript{th} resource on round $t$.  
%As a motivational example, the edges could denote wireless links and the vector $x_i(t)$ may encode the ON/OFF status of the set of links incident to the $i$\textsuperscript{th} agent. %Similarly, 
Let the binary action variable $y_{ij}(t)\in [0,1]$ denote the amount by which the agent $i$ is matched with resource $j$ on round $t$. Hence, the reward accrued by the agent $i$ on time $t$ is given by $\langle x_i(t), y_i(t) \rangle.$ Let $\Delta$ denote the convex hull of all matchings. It is well known that $\Delta$ can be succinctly represented by the set of all $m\times m$ doubly stochastic matrices, \emph{a.k.a.} the Birkhoff polytope \citep{ziegler2012lectures}). In other words, in the \texttt{OFM} problem, the set of all feasible actions $\Delta$ consists of all $m\times m$ matrices $\big(y_{ij}\big)_{i,j}$ satisfying the following constraints \footnote{As before the fractional matching can be converted to a randomized integral matching via sampling. See section \ref{int-alloc} for details.}: 
\begin{eqnarray} \label{bvn-polytope}
	\sum_{i=1}^m y_{ij}=1,~ \forall j,~~
	\sum_{j=1}^m y_{ij}=1,~ \forall i,~~
	0\leq y_{ij} \leq  1, ~ \forall i,j.
\end{eqnarray} 

Let the variable $R_i(t)$ denote the cumulative rewards accried by the $i$\textsuperscript{th} agent up to round $t $. Clearly,
\begin{eqnarray} \label{reward-evolution}
	R_i(t+1) = R_i(t) + \langle x_i(t), y_i(t) \rangle, R_i(0)=0, ~~1 \leq i \leq m,
\end{eqnarray}
%\Cam{Observe here that assumption 2 holds?}
It can be verified that Assumption 2 holds in this problem with $\mu=m^{-1}$ by noting that $m^{-1} \bm{1}_{m\times m} \in \Delta.$
 The objective of the Online Fair Matching (\texttt{OFM}) problem is to design an online matching policy that minimizes the $c$-Regret \eqref{regret-def}. Figure \ref{ofm} illustrates a special case of the problem for binary-valued demands. From the above formulation, it can be immediately seen that the Online Matching problem is an instance of the \texttt{NOFRA} problem. Furthermore, it also generalizes the online scheduling problem described in Example \ref{sched-example}.

The online matching problem arises in numerous practical settings. %We list below a few common scenarios where one needs to solve an online matching problem efficiently. 
 For example, in the problem of online Ad Allocation, there are $m$ different advertisers whose ads need to be placed in $m$ different display slots on a webpage. Each slot can accommodate only one ad. On round $t$, a new user arrives and presents a reward vector $\bm{x}_i(t)$ for each advertiser. For example, the component $x_{ij}(t)$ could denote potential revenue accrued by the advertiser if the ad $i$ is placed on the $j$\textsuperscript{th} slot on round $t$. The objective of the allocation policy is to match the ads to the slots on each round so that the total earned revenue is fairly distributed among the advertisers. Similar problems arise in designing recommendation systems for crowdsourcing or online dating websites \citep{tu2014online}, fair channel assignment in wireless networks \citep{ALTMAN2010338}. 
	%\item  In the context of wireless networks, assume that the agents represent the users and the resources represent different wireless channels (frequencies). Due to interference constraints, each user can be allotted only one channel, hence the allocation must be a matching. The reward vector $x_i(t)$ denotes the gains of the $i$\textsuperscript{th} user for different channels on round $t$. The gains are revealed only after the end of each rounds.  
%	\end{enumerate}

%\textbf{\edit{Verify:} }It is easy to ensure that Assumptions 1 and 2 hold for the above two problems. 
\begin{figure}[!tbp]
  \centering
  \begin{minipage}[b]{0.3\textwidth}
   \centering
    \includegraphics[scale=0.3]{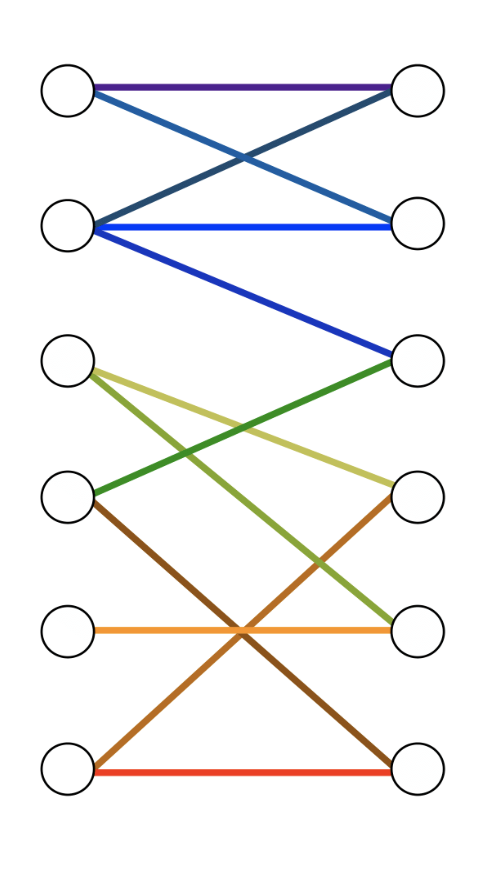}
    \put(-100,60){Agents}
    \put(-4,60){Resources}
   \caption*{A Bipartite graph $\mathcal{G}$}
  \end{minipage}
  %\hfill
  \begin{minipage}[b]{0.3\textwidth}
   \centering
    \includegraphics[scale=0.3]{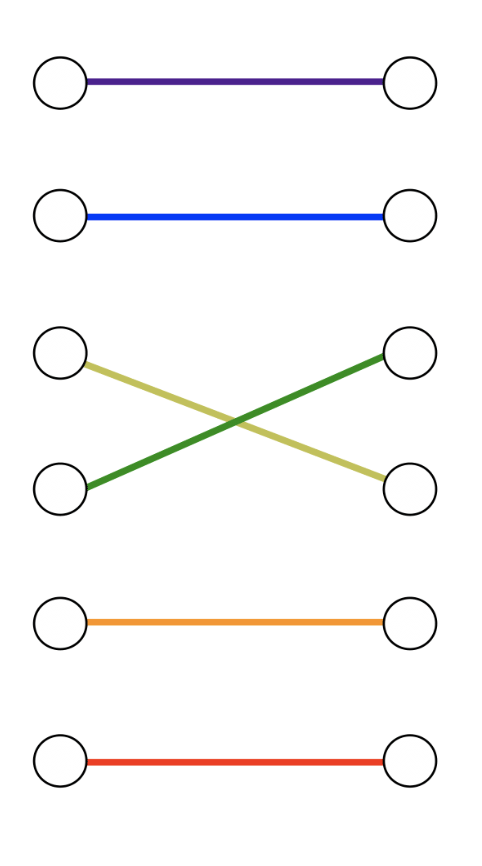}
   \caption*{A Matching in $\mathcal{G}$}
  \end{minipage}
  \caption{Illustrating the Online Fair Matching (\texttt{OFM}) problem for zero-one demands. The existence of an edge denotes unit demand, and likewise, the absence of any edge implies zero demand.}
  \label{ofm}
\end{figure}
\end{example}
%\paragraph{Example 4: Online Job Scheduling \citep{even2009online}:} 

\section{Designing an approximately-no-regret policy for \texttt{NOFRA}} \label{analysis}
In view of the difficulty outlined in Section \ref{difficulty}, the design and analysis of the Online Proportional Fair (\texttt{OPF}) policy consists of two parts. First, in Lemma \ref{regret-comp}, we show that by greedily estimating the terminal gradient with the current gradients at each round, the standard regret of the \texttt{NOFRA} problem can be upper bounded by the $(1-\alpha)^{-(1-\alpha)}$-approximate static regret of a surrogate online linear optimization problem with a policy-dependent gradient sequence. Finally, we use the online gradient ascent policy with adaptive step sizes to solve the surrogate problem while simultaneously controlling the magnitude of the gradients. The following section details our technique. 
%The central idea behind the design of our online policy is to reduce the \texttt{NOFRA} problem to an instance of surrogate online linear optimization problem.
\subsection{Reducing the \texttt{NOFRA} problem to an Online Linear Optimization Problem with policy-dependent subgradients}

%, which can be solved with the standard OCO techniques \citep{SIGMETRICS20, mukhopadhyay2022k}. 
%In doing so, it turns out that one also needs to slightly perturb the input request sequence itself to make the reduction work.
% Recall that, for any differentiable concave function $\psi: \mathcal{X} \to \mathbb{R},$ we have the following bound from its first-order Taylor expansion:
%\begin{eqnarray} 
%	\psi(z_1) - \psi(z_2) \leq \langle \nabla \psi (z_2), z_1-z_2 \rangle, ~~\forall z_1, z_2 \in \mathcal{X}. 
%\end{eqnarray}
%

Since the fairness function $\phi: \mathbb{R}_+ \to \mathbb{R}_+$ is concave, we have that for any $x,y \geq 0:$
\begin{eqnarray}\label{conc-taylor}
	\phi(x) - \phi(y) \leq \phi'(y)(x-y).
\end{eqnarray} 
Let $\beta \geq 1$ be some constant to be determined later. Taking $x=R_i^*(T)$ and $y=\beta R_i(T)$ in the above inequality, we have 
%of the previous action variables $\{y_i(\tau)\}_{\tau=1}^T$ only, using the first-order Taylor series bound for concave functions between the points $\{y^*_i\}_{t=1}^T$ and $\{\beta y_i(t)\}_{t=1}^T$, we can write 
\begin{eqnarray} \label{first-order-bound}
\phi(R_i^*(T))- \beta^{1-\alpha}\phi(R_i(T))
&\stackrel{(a)}{=}&	\phi(R_i^*(T)) - \phi(\beta R_i(T)) \nonumber, \\
&\stackrel{(b)}{\leq} & \phi'(\beta R_i(T))(R_i^*(T) - \beta R_i(T)) \nonumber \\
&\stackrel{(c)}{=}&  \beta^{-\alpha}\phi'(R_i(T)) \sum_{t=1}^T \langle x_i(t), y_i^*-\beta y_i(t) \rangle.  
\end{eqnarray}
where in (a), we have used the positive homogeneity property of the fairness function $ \phi(\beta x) = \beta^{1-\alpha} \phi(x),$ in (b), we have used the concavity of the fairness function $\phi(\cdot)$ from Eqn.\ \eqref{conc-taylor}, and, finally, in (c), we have used the definition of cumulative rewards \eqref{hit-eqn} and the homogeneity of the function $\phi(\cdot)$ to conclude that $\phi'(\beta x)= \beta^{-\alpha} \phi'(x).$ 
%
%explicit form of the $\alpha$-fair utility function to conclude that \footnote{We need to check if this type of analysis extends to positively homogeneous reward functions, such as norms. A function $f: \mathcal{X} \to \mathbb{R}$ is positively homogeneous if $f(\beta x) = |\beta| f(x)$ for $x \in \mathcal{X} \subset \mathbb{R}^d, \beta >0.$}
%\[ \phi(\beta x) = \beta^{1-\alpha} \phi(x), ~~\phi'(\beta x) = \beta^{-\alpha} \phi'(x).\]
%In the equality (a), we have also exploited the homogeneity of the cumulative hits $R_i(T)$ w.r.t. the action variables $\{y_i(t)\}_{t=1}^T$ (Eqn.\ \eqref{hit-eqn}) to conclude that scaling all action variables by a factor $\beta$ scales the cumulative hits by the same factor.  

Summing up the bound \eqref{first-order-bound} over all agents $i \in [m]$, we obtain the following upper bound to the $\beta^{1-\alpha}$-Regret of any online policy for the \texttt{NOFRA} problem:  
\begin{eqnarray} \label{regret-expr3}
	\textrm{Regret}_T(\beta^{1-\alpha}) \leq \beta^{-\alpha}\sum_{t=1}^T \bigg( \sum_{i} \langle \phi'(R_i(T)) x_i(t), y_i^*-\beta y_i(t) \rangle \bigg).
\end{eqnarray}

It is critical to note that each of the quantities $\phi'(R_i(T))$ depends on the entire sequence of past actions $\{y_t\}_{t=1}^T.$ Hence, \eqref{regret-expr3} is \emph{not} the standard regret as the terminal value of $\phi'(R_i(T))$ is not known to the online policy when it takes its actions. More importantly, the value of the coefficient $\phi'(R_i(T))$ depends on the future actions and requests through \eqref{recursion}, and hence, it is \emph{impossible} to know the value of this coefficient in an online fashion. To get around this fundamental difficulty, we now consider a \emph{surrogate} regret-minimization problem by replacing the term $\phi'(R_i(T))$ by its \emph{causal version} $\phi'(R_i(t))$ on round $t$. 
%and setting $\beta=1,$
In other words, we now seek to design an online learning policy $\pi$ that minimizes the regret of the following surrogate online linear optimization problem:
 \begin{eqnarray} \label{surr-regret}
 	\hat{\textrm{Regret}_T} \equiv  \sum_{t=1}^T \bigg( \sum_{i}\langle \phi'(R_i(t)) x_i(t), y_i^*- y_i(t) \rangle \bigg).
 \end{eqnarray}
To recapitulate the information structure, recall that the cumulative reward vector $\bm{R}(t)$ is known, but the current request vector $\bm{x}(t)$ is unknown to the online learner before it makes a decision $y_t$ on the $t$\textsuperscript{th} round. We now establish the following central result that relates the bound \eqref{regret-expr3} to the regret bound \eqref{surr-regret} of the online linear optimization problem. 
%which is central to the design and analysis of our policy: 
\begin{framed}
\begin{lemma} \label{regret-comp}
Consider any arbitrary online resource allocation policy and assume that the utility function of each user is given by the $\alpha$-fair utility function \eqref{alpha-fair}. Then, we have
%for any fixed benchmark caching configuration $y^*$, we have
\begin{eqnarray} \label{regret-comp-eq}
	\textrm{Regret}_T(c_\alpha) \leq (1-\alpha)^{\alpha}\hat{\textrm{Regret}_T},
\end{eqnarray}
 where $c_\alpha \equiv (1-\alpha)^{-(1-\alpha)} \leq e^{1/e}<1.445.$
\end{lemma}
\end{framed}
%\edit{Include proof ideas}.
\paragraph{Proof outline}
As discussed above, the surrogate problem greedily replaces the non-causal terminal gradient $\phi'(R_i(T))$ in Eqn.\ \eqref{regret-expr3} by the current gradient $\phi'(R_i(t)), \forall i$ on round $t$. The proof of Lemma \ref{regret-comp} revolves around showing that this transformation can be done by incurring only a small penalty factor to the overall regret bound. For proving this result, we split the regret expression \eqref{regret-expr3} into the difference between two terms -  term $(A)$ corresponding to the cumulative reward accrued by the benchmark static allocation ($\bm{y}^*$), and term $(B)$ corresponding to the reward accrued by the online policy. Next, we compare these two terms separately with the corresponding terms $(A')$ and $(B')$ in the regret expression for the surrogate online linear optimization problem \eqref{surr-regret}. By exploiting the non-decreasing nature of the cumulative rewards, we first show that $(A) \leq (A')$ (see Eqn.\ \eqref{bd1}). Next, by using form of the $\alpha$-fair utility function, we show that $(B') \leq (1-\alpha)^{-1}(B)$ under the action of any policy (see Eqn.\ \eqref{bd2}). Lemma \ref{regret-comp} then follows by combining the above two results.    
See Appendix \ref{regret-comp-proof} for the proof of Lemma \ref{regret-comp}. 

\subsection{Online Learning Policy for the Surrogate Problem and its Regret Analysis}

  \begin{algorithm}
\caption{The Online Proportional Fair (\texttt{OPF}) Policy}
\label{OPFC-main}
\begin{algorithmic}[1]
\State \algorithmicrequire{ Fairness parameter $0\leq \alpha<1$,} Demand/Reward vectors from the agents $\{\bm{x}(t)\}_{t=1}^T,$ Euclidean projection oracle $\Pi_{\Delta}(\cdot)$ onto the feasible set $\Delta,$ an upper-bound $D$ to the Euclidean diameter of the feasible set\footnotemark.
\State \algorithmicensure{ Online resource allocation decisions $\{y_t\}_{t=1}^T$ }
\State $R_i \gets 1, \forall i\in [m], \bm{y}\gets \mu\mathds{1}_{N\times m}, S\gets 0$ \algorithmiccomment{\textcolor{black}{\emph{Initialization}}}
\ForEach {round $t=2:T$:}
\State $g_i \gets  \frac{x_i(t-1)}{R_i^\alpha}, \forall i \in [m]$ \algorithmiccomment{\emph{Computing the gradient components for each agents}} \label{pf-step}
\State $g\gets (g_1, g_2, \ldots, g_m)$  \algorithmiccomment{\emph{Computing the full gradient}}
\State $S \gets S+ ||g||_2^2,$   \algorithmiccomment{\emph{Accumulating the norm of the gradients}}
\State $y \gets \Pi_{\Delta}\big(y+ \frac{D}{2\sqrt{S}} g\big)$ \algorithmiccomment{\textcolor{black}{\emph{Updating the inclusion probabilities using OGA}}} \label{proj}
\State [\textbf{Optional}] Sample a randomized integral allocation $Y$ s.t. $\mathbb{E}[Y]=y.$ \label{sampling-step}
\State The agents reveal their demand/ reward vectors $\{x_i(t)\}_{i \in [m]}$ for the current round.
\State $R_i \gets R_i + \langle x_i(t), y_i(t) \rangle, ~\forall i\in [m].$ \algorithmiccomment{\textcolor{black}{\emph{Updating cumulative rewards}}} \label{updates}
%\State Cache file $S_t \gets \textsc{Madow-Sample}(y)$
%$k$ files according to the inclusion probability vector $y$ using Madow's sampling scheme described in Algorithm \ref{madow}
%given in Algorithm \ref{uneq} 
%\algorithmiccomment{\textcolor{black}{\emph{Sample a subset of $k$ files by invoking Algorithm \ref{Madow}}}}
%according to the \textsc{FTRL}($\eta$) update 
%\eqref{ftrl-opt}. 
%\STATE Feed the vector $\bm{g}_t$ as the reward vector for round $t$ to the \textsc{FTRL} policy 
\EndForEach
\end{algorithmic}
\end{algorithm}
\footnotetext{See Appendix \ref{dia-bd} for upper bounds on the Euclidean diameter of the action sets for Examples 2.1-2.3.}

Having established that the $c$-regret of the original problem is upper bounded by the regret of the surrogate learning problem, we now proceed to upper bound the regret of the surrogate problem under the action of a \emph{no-regret} learner. In the sequel, we use the projected Online Gradient Ascent (OGA) with adaptive step sizes \citep[Algorithm 2.2]{orabona2019modern} for designing a no-regret policy for the surrogate problem. We call the resulting online learning policy \textsc{Online Proportional Fair} (\texttt{OPF}). The pseudocode of the \texttt{OPF} policy is given in Algorithm \ref{OPFC}.

%The term \emph{Proportional Fairness} formalizes the intuition that the gradient component of any user in the online policy is discounted by a factor proportional to its current cumulative reward $R_i^\alpha(t)$ (step \ref{pf-step} of Algorithm \ref{OPFC}). 
\paragraph{The Online Proportional Fair policy (\texttt{OPF}):}
At the end of each round, each user $i$ computes its gradient component $g_i$ by dividing its current demand vector $x_i(t-1)$ with its current cumulative reward $R_i(t)$ raised to the power $\alpha$ (line \ref{pf-step} of Algorithm \ref{OPFC-main}). This formalizes the intuition that a user with a larger current cumulative reward has a smaller gradient component. In line \ref{proj}, we take a projected gradient ascent step, where the projection operation constitutes the main computational bottleneck. In practice, the projection can often be computed efficiently using variants of the Frank-Wolfe algorithm while given access to an efficient LP oracle. 
 With additional structure, the \texttt{OPF} policy can be simplified further with a more efficient projection. For example, we give a simplified implementation for the online shared caching problem in Appendix \ref{simplified} by exploiting the fact that the action vector $y(t)$ is the same for all users, \emph{i.e.,}
 %following structure of the action space: 
 $y_1(t)=y_2(t)=\ldots =y_m(t)\equiv y(t), \forall t.$ Step \ref{sampling-step} is an optional sampling step which is executed only if an integral allocation is required. For pedagogical reasons, we will skip Step \ref{sampling-step} in this section and discuss it later in Section \ref{int-alloc}. Finally, the cumulative rewards of all users are updated in step \ref{updates}. The following lemma gives an upper bound to the regret of the \texttt{OPF} policy for the surrogate problem. 
%\newpage
%request of the $i$\textsuperscript{th} user is weighted down by a factor of $R_i^\alpha(t)$.  
%In designing the \texttt{OPFC} policy, we have used the fact that the Euclidean diameter of the feasible set $\Delta^N_k$ is bounded by $\sqrt{2k}$ (shown in Lemma \ref{dia-bound}). The pseudocode for this policy is shown in Algorithm \ref{OPFC}.
%\edit{Explain the algorithm briefly}
% 
%\paragraph{Technical insight:}
%Note that a tight regret analysis of the \texttt{OPF} policy for the surrogate problem is highly non-trivial as the norms of the future gradients are implicitly controlled by its past actions $\{\bm{y}(\tau)\}_{\tau=1}^t$  that control the future cumulative rewards $\bm{R}(t)$. This is in sharp contrast with the standard online learning problem where the upper bound to the norm of the gradients remains fixed \emph{a priori} for all admissible policies. Hence, a tight regret analysis should exploit the fact that, under the \texttt{OPF} policy, the norm of the gradients diminishes at an appropriate rate. In Theorem \ref{reg-bd-alpha}, we derive a tight regret bound for the surrogate problem. The proof makes use of a new \emph{Bootstrapping} technique, illustrated in Figure \ref{bootstrap}, which may be of independent interest.
% We now prove the following regret bound for the surrogate problem. 
%\section{Regret analysis of the Surrogate Problem} \label{hit-rate-lb}
\begin{framed}
\begin{lemma} \label{reg-bd-alpha}
	The Online Proportional Fair policy, described in Algorithm \ref{OPFC-main} achieves the following standard regret bound for the surrogate problem \eqref{surr-regret} for the $\alpha$-fair utility function:
	\begin{eqnarray*}
	\hat{\textrm{Regret}}_T  = \begin{cases}
	O(T^{1/2-\alpha}), ~~ \textrm{if} ~~0<\alpha <\nicefrac{1}{2}, \\
		O(\sqrt{\log T}),~~ \textrm{if} ~~\alpha=\nicefrac{1}{2}\\
		O(1), ~~ \textrm{if} ~~\nicefrac{1}{2}<\alpha <1.
	\end{cases}
\end{eqnarray*}
	
	 Furthermore, under this policy, the cumulative rewards of each user increases linearly with time, \emph{i.e.,} $R_i(T)=\Omega(T), \forall i, T$.  
	%and ensures that $R_i(t) \geq \Omega(\frac{t}{\log t}), \forall i, t.$
\end{lemma}
\end{framed}
\paragraph{Proof outline:}
One of the major challenges in the regret analysis of the surrogate problem \eqref{surr-regret} is that the coefficients of the gradients $\{\phi'(R_i(t)), i\in [m]\}$ on round $t$ depends on the past actions $\{\bm{y}(\tau)\}_{\tau=1}^t$ of the policy itself. Since the regret bound of any online linear optimization problem scales with the norm of the gradients, we now need to \emph{simultaneously} control the regret \emph{and} the norm of the gradients generated by the online policy. Surprisingly, the proof of Theorem \ref{reg-bd-alpha-proof} shows that the proposed online gradient ascent policy with adaptive step sizes not only provides a sublinear regret but it also keeps the gradients small, which in turn helps keep the regret small. In fact, these two goals are well-aligned and our proof precisely exploits the reinforcing nature of these two objectives via a new \emph{Bootstrapping} technique. See Appendix \ref{reg-bd-alpha-proof} for the proof of Lemma \ref{reg-bd-alpha}. 
Finally, combining Lemma \ref{regret-comp} and Lemma \ref{reg-bd-alpha}, we obtain the main result of this paper.
%yields a sublinear approximate-regret bound for the \texttt{NOFRA} problem. 
\begin{framed}
\begin{theorem}[(Regret bound of the \texttt{OPF} policy)] \label{main-result}
	The Online Proportional Fair (\texttt{OPF}) policy, described in Algorithm \ref{OPFC-main}, achieves  the following approximate regret bound for the \texttt{NOFRA} problem \eqref{regret-def}:
	\begin{eqnarray} \label{main-thm-bd}
	\textrm{Regret}_T(c_\alpha)  = (1-\alpha)^\alpha \begin{cases}
	O(T^{1/2-\alpha}) ~~ \textrm{if} ~~0<\alpha <\nicefrac{1}{2}, \\
		O(\sqrt{\log T})~~ \textrm{if} ~~\alpha=\nicefrac{1}{2}\\
		O(1) ~~ \textrm{if} ~~\nicefrac{1}{2}<\alpha <1,
	\end{cases} 	\end{eqnarray}
 where $c_\alpha\equiv (1-\alpha)^{-(1-\alpha)}$.
\end{theorem}	
\end{framed}
 \paragraph{Remarks:} 
 
1. For the job scheduling problem in the reward maximization setting, \citet[Lemma 4]{even2009online} showed that if the cumulative reward is concave and the offline optimal reward is convex, then their proposed (albeit complex) approachability-based recursive policy achieves zero-regret. In Section 7 of the same paper, the authors posed an open problem of attaining a relaxed goal when the above sufficient condition is violated. In Appendix \ref{non-cvx}, we show that for the $\alpha$-fair utility function \eqref{alpha-fair}, 
 % and the $(1-\alpha)$-norm reward function \eqref{norm-def}, 
 the offline optimal cumulative reward is \emph{non-convex} in the regime $0<\alpha <1$. Hence, Theorem \ref{main-result} gives a resolution to the above open problem by exhibiting a simple online policy with a sublinear approximate regret when the given sufficient condition is violated. Furthermore, the computational complexity of our policy is linear in the number of machines $m,$ which is way better than their approachability-based policy whose complexity scales exponentially in $m$.
 
%\textbf{Note:} 

2. Observe that the regret bound given by Theorem \ref{reg-bd-alpha} always remains non-vacuous, \emph{irrespective} of the value of the fairness parameter $\alpha$ and the sequence of adversarial reward vectors. This follows from the fact that irrespective of the demand vectors, by choosing the constant action $y=\mu\mathds{1},$ each user can achieve a cumulative reward of $\mu \delta T.$ Hence, the optimal offline value of the $\alpha$-fair utility function $\Omega(T^{1-\alpha}).$ On the other hand, the \texttt{OPF} policy achieves a regret bound of $O(T^{\nicefrac{1}{2}-\alpha}),$ (Theorem \ref{main-result}) that is always dominated by the optimal static offline objective.

%2. It is not surprising that the policy \texttt{OPF} achieves a constant regret in the regime $1/2 < \alpha <1.$ It can be easily verified that, for $\alpha=1$ (which corresponds to the logarithmic utility function), the oblivious randomized policy, which chooses an action uniformly at random, achieves constant regret. 

3. In the special case of $\alpha=0,$ the \texttt{NOFRA} problem corresponds to the cumulative reward maximization problem for all users. Hence, Theorem \ref{reg-bd-alpha} recovers the well-known $O(\sqrt{T})$ standard regret bound obtained by \cite{SIGMETRICS20} in the context of online caching.

The following converse result gives a universal lower bound to the approximation factor $c$ for which it is possible to design an online policy with a sublinear $c$-regret.
%\newpage
\begin{framed}
\begin{theorem}[(Lower bound to the approximation factor)] \label{lb-thm}
	Consider the online shared caching problem for the $\alpha$-fair reward function with $m=2$ users. Then for any online policy with a sublinear $c_\alpha$-regret, we must have \[c_\alpha \geq \max_{0 \leq \eta \leq \nicefrac{1}{2}}\frac{{\eta }^{1-\alpha} + {(1-\eta)}^{1-\alpha}}{{(1-\eta/2) }^{1-\alpha} + {({\eta }/{2})}^{1-\alpha}}>1, ~~ 0< \alpha <1.\]
	%It is impossible to design a no-$c$-approximate-regret policy for this problem with $c<1.06.$ 
\end{theorem}
\end{framed}
A numerical comparison between the upper and lower bounds on the approximation factor is shown in Figure \ref{fig:c-lower-bound}. See Appendix \ref{lb-proof} for the proof of Theorem \ref{lb-thm}.

\begin{figure}[t ]
    \centering
    \includegraphics[scale=0.42]{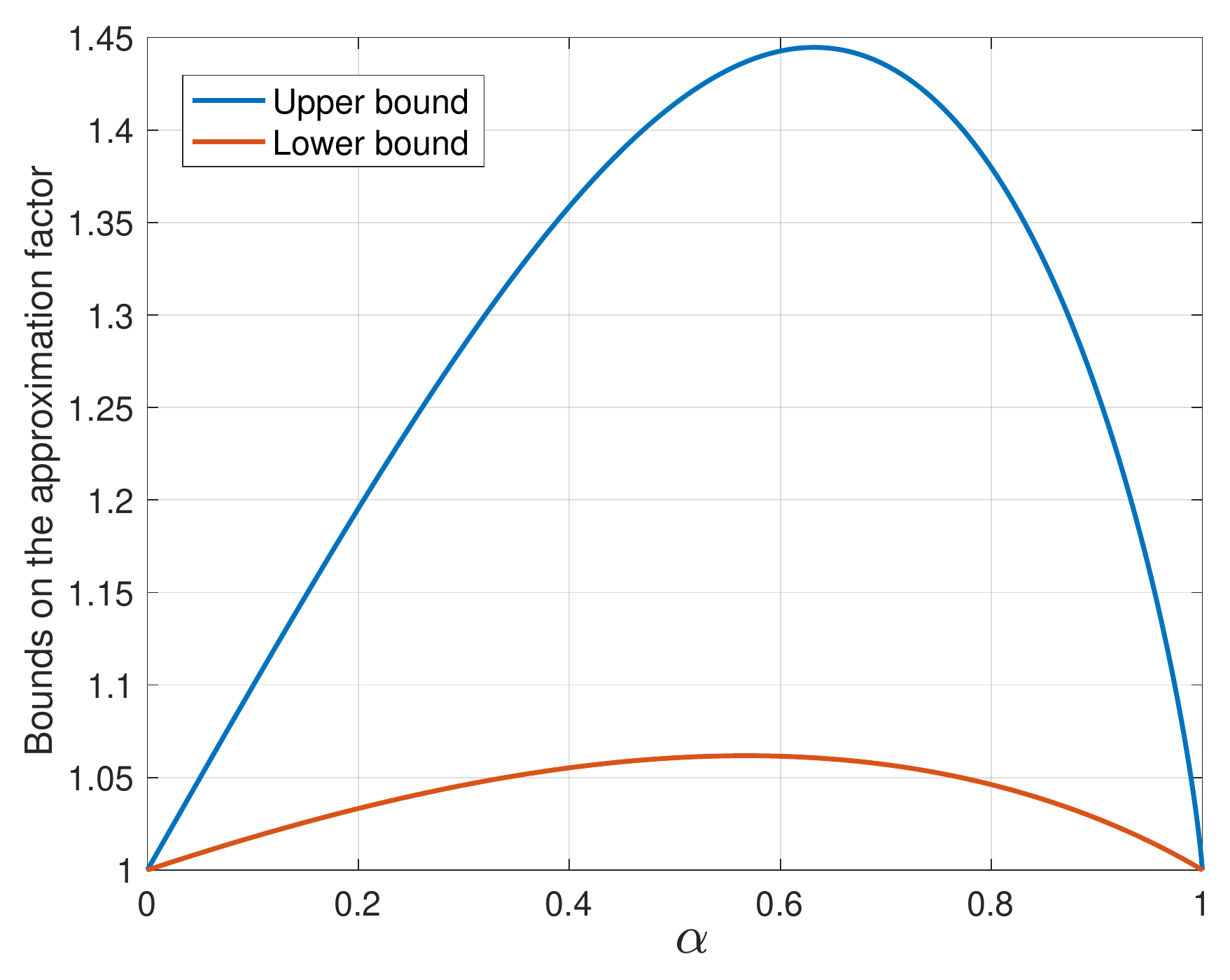}
    \put(-113,130){\small{$(1-\alpha)^{-(1-\alpha)}$}}
    \caption{Comparison between the upper and lower bounds of the approximation ratio}
    \label{fig:c-lower-bound}
\end{figure}

\section{High-probability regret bound for randomized integral allocations} \label{int-alloc}
Theorem \ref{reg-bd-alpha} establishes $c_\alpha\equiv (1-\alpha)^{-(1-\alpha)}$-regret guarantee for the Online Proportional Fair (\texttt{OPF}) policy for fractional allocations where the set of admissible actions $\Delta$ is the convex hull of integral allocations. In the following, we show that a sub-linear $c_\alpha$-regret guarantee holds with high probability when randomized integral actions are chosen according to the optional step (\ref{sampling-step}) in the \texttt{OPF} policy. Recall that for integral allocation, on each round $t$ we independently sample an integral allocation $Y_t \in \{0,1\}^{mN},$ that matches with the fractional allocation $y_t$ in expectation, \emph{i.e.,} $\mathbb{E}[Y_t] = y_t$ where $y_t$ is the fractional allocation vector recommended by the \texttt{OPF} policy. A fractional allocation vector $y_t$ in the convex hull of integral allocations can be turned to a randomized integral allocation by expressing $y_t$ as a convex combination of integral allocations and then randomly sampling one of the integral allocations with appropriate probabilities. As Appendix \ref{sampling-tech} shows, for many problems with structure, such a decomposition can be efficiently computed. 
%Using the standard concentration bounds, we now show that with high probability the randomized integral allocation policy achieves similar regret guarantee as the deterministic fractional allocation policy.   
\paragraph{Regret analysis:}
By appealing to \citep[Lemma 4.1]{cesa2006prediction}, it is enough to consider an oblivious adversary that fixes the sequence of demand vectors $\{\bm{x}(t)\}_{t\geq 1}$ before the game commences. 
Let the random variable $\mathsf{R}_i(T)$ denote the random cumulative reward obtained by the $i$\textsuperscript{th} user under the randomized integral allocation policy and the deterministic variable $R_i(T)$ is defined as in Eqn.\ \eqref{hit-eqn}. By construction, we have $\mathbb{E}[\mathsf{R}_i(T)] = R_i(T).$ Furthermore, since $\mathsf{R}_i(T)$ is the sum of $T$ independent random variables, each of magnitude at most one, from the standard Hoeffding's inequality, we have 
\begin{eqnarray*}
		\mathbb{P}(|\mathsf{R}_i(T) - R_i(T)| \geq \lambda ) \leq \exp(-2\lambda^2/T), \forall i \in [m].  
	\end{eqnarray*}  
	
	Hence, with probability at least $1-\texttt{poly}(1/T),$ the aggregate $\alpha$-fair utility \eqref{alpha-fair}
	 accrued by the randomized policy can be lower bounded as:
	\begin{eqnarray} \label{conc-ineq1}
	 (1-\alpha)^{-1}\sum_i\mathsf{R}^{1-\alpha}_i(T) &\geq&   (1-\alpha)^{-1} \sum_i \bigg(R_i(T) - O(\sqrt{T \log T})\bigg)^{1-\alpha} \nonumber \\
		 &\stackrel{(a)}{\geq} &  (1-\alpha)^{-1}\sum_i R_i^{1-\alpha}(T) - O\big((T \log T)^{\frac{1-\alpha}{2}}\big),
	\end{eqnarray}
	where in inequality (a), we have used the fact that $(x+y)^{1-\alpha} \leq x^{1-\alpha}+y^{1-\alpha}, \forall 0\leq \alpha \leq 1, x,y\geq 0.$ Combining the above with the approximate regret bound in Theorem \ref{main-result}, and taking the dominant term, we conclude that the $c_\alpha$-regret for the randomized integral allocation policy is upper-bounded by $ O\big((T \log T)^{\frac{1-\alpha}{2}}\big)$ w.h.p. for all $0\leq \alpha \leq 1.$ Unfortunately, unlike Theorem \ref{main-result}, the integral allocation policy incurs a non-trivial (but sublinear) $c_\alpha$-regret for the entire range of the fairness parameter $0\leq \alpha <1.$

\section{Conclusion and open problems} \label{conclusion}
In this paper, we propose an efficient online resource allocation policy, called Online Proportional Fair (\texttt{OPF}), that achieves a $c_\alpha$-approximate sublinear regret bound for the $\alpha$-fairness objective, where $c_\alpha \equiv (1-\alpha)^{-(1-\alpha)} \leq 1.445,$ for  $0< \alpha <1.$ Our main technical contribution is to show that the non-additive $\alpha$-fairness function can be efficiently learned by greedily estimating the terminal gradients. An important follow-up problem is to investigate the extent to which the algorithmic and analytical methodologies introduced in this paper can be generalized. Specifically, it would be interesting to see if a similar online policy can be designed for the online scheduling problem in the regime $\alpha >1.$ Note that, for the online scheduling problem, only a recursive approachability-based policy is known in the literature, whose complexity scales exponentially with the number of machines \citep{even2009online}. Another related problem is to design an optimistic version of the proposed \texttt{OPF} policy that offers an improved regret bound by efficiently incorporating hints regarding the future demand sequence \citep{mhaisen2022optimistic, bhaskara2020online}. Furthermore, reducing the gap between the upper and lower bounds of the approximation factor in Figure \ref{fig:c-lower-bound} would be interesting. Finally, designing an online resource allocation policy with a small dynamic regret would give a more fine-grained regret bound depending on the regularity of the demand/reward sequence.  
%\begin{enumerate}
%\item Prove expected regret bounds in the discrete caching set up
	%\item Proving tight lower bounds on regret for different values of $\alpha$ (Done!)
%	\item Using predictions for the next file requests to reduce the regret bound
	%\item Designing policies with a small dynamic regret
	%\item Approach using adversarial RL
	%\item Investigating what happens when the users intermittently come and go. If the user $i$ is not present on round $t$, it simply requests the zero vector.
	%\item Bounds on Dynamic Regret
%\end{enumerate}
\section{Acknowledgement}
This work was supported by a US-India NSF-DST collaborative grant coordinated by IDEAS-Technology Innovation Hub (TIH) at the Indian Statistical Institute, Kolkata. C. Musco was also partially supported by an Adobe Research grant.

%\newpage

%\clearpage
%\input{extensions}
%\input{analysis}
\section{Appendix}
\subsection{Proof of the inequality $\phi'(R_i(t)) (R_i(t+1)-R_i(t)) \leq \int_{R_i(t)-1}^{R_i(t+1)-1} \phi'(R) dR$} \label{ineq-comp}
We use the fact that $\phi'(\cdot)$ is a non-increasing function and $R_i(t+1)-R_i(t) \leq 1.$ The following diagram is self-explanatory.
\begin{figure}[H]
\centering
	\includegraphics[scale=0.6]{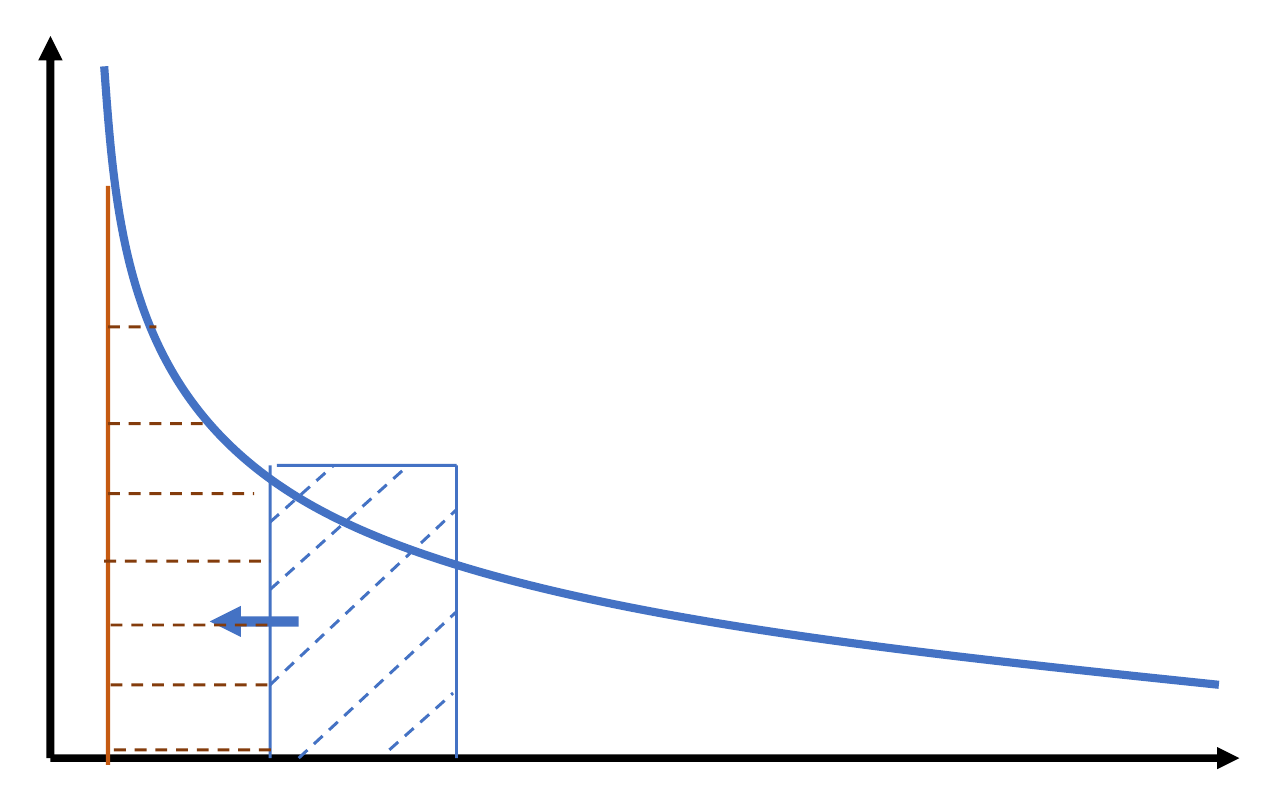}
	\put(-160,20){\footnotesize{$\phi'(R_i(t)) (R_i(t+1)-R_i(t))$}}
	\put (-185, 70){\footnotesize{$\phi'(R)$}}
	\put(-170, 5){$\underbrace{\textrm{\hspace{25pt}}}$}
	\put(-142,-2){ $\textrm{length}\leq 1$}
	\put(-190,90){$\int_{R_i(t)-1}^{R_i(t+1)-1} \phi'(R) dR$}
	\put(-70, 100){\footnotesize{$ R_i(t+1)-R_i(t) \leq 1$}}
	\put(-140, 50){\small{Translate the rectangle to the left by one unit} }
	\caption{Graphical proof of the upper bound}
\end{figure}

\subsection{Upper bound on the diameter of the admissible sets} \label{dia-bd}
\begin{lemma} \label{dia-bound-ofc}
	For the online shared caching  problem the diameter of the admissible action set can be bounded as follows: \[\textrm{Diam}(\Delta^{N}_k) \leq \sqrt{2k}.\]
\end{lemma}

\begin{proof}
	Let $x, y \in \Delta^N_k.$ We have 
	\begin{eqnarray*}
		||x-y||_2^2 = \sum_{i=1}^N (x_i-y_i)^2 = \sum_{i=1}^N |x_i-y_i| |x_i-y_i| \stackrel{(a)}{\leq} \sum_{i=1}^N |x_i-y_i|\stackrel{(b)}{\leq} \sum_{i=1}^N |x_i| + \sum_{i=1}^N|y_i| \stackrel{(c)}{=} 2k,
	\end{eqnarray*}
	where, in (a), we have used the fact that $0\leq x_i, y_i \leq 1,$ in (b), we have used triangle inequality, and in (c), we have used the fact that the sum of the components of each feasible vector is $k$.
\end{proof}

\begin{lemma} \label{dia-bound-ofm}
	For the online matching problem, the diameter of the admissible action set can be bounded as follows:\[\textrm{Diam}(\Delta) \leq \sqrt{2m}.\]
\end{lemma}
	Let $x, y \in \Delta,$ where $\Delta$ is the feasible set for the \texttt{OFM} problem. We have 
	\begin{eqnarray*}
		||x-y||_2^2 = \sum_{i,j}(x_{ij}-y_{ij})^2 \leq \sum_{i,j}|x_{ij}-y_{ij}| \leq \sum_{ij}x_{ij} + \sum_{ij}y_{ij} \leq 2m,
		\end{eqnarray*}
where the inequalities follow from similar arguments as in the proof of the previous lemma. 

\subsection{Proof of Lemma \ref{regret-comp}} \label{regret-comp-proof}

\begin{proof}
%It is sufficient to consider the case when the $c_\alpha-\textrm{Regret}_T$ in the LHS of \eqref{regret-comp-eq} is non-negative. In case the $c_\alpha-\textrm{Regret}_T$ is negative for some benchmark $y^*$, there is nothing to prove.

The expression for $\textrm{Regret}_T(\beta^{1-\alpha})$ from Eqn.\ \eqref{regret-expr3} can be split into two terms as follows:
	\begin{eqnarray*}
		\textrm{Regret}_T(\beta^{1-\alpha}) =\beta^{-\alpha}\big[\underbrace{\sum_i \phi'(R_i(T)) \sum_{t=1}^T \langle x_i(t), y_i^* \rangle}_{\textrm{\textcolor{blue}{$(A)$}}} - \beta \underbrace{\sum_i \phi'(R_i(T)) \sum_{t=1}^T \langle x_i(t), y_i(t) \rangle}_{\textrm{\textcolor{blue}{$(B)$}}}\big].  
	\end{eqnarray*}
	Also denote the corresponding terms in the regret expression \eqref{surr-regret} for the surrogate learning problem by $(A')$ and $(B')$. 
	We will now separately bound each of the above two terms in terms of the corresponding terms in the regret expression \eqref{surr-regret} for the surrogate learning problem. 
	
	\paragraph{\underline{Proving \textcolor{blue}{$(A)\leq (A')$}}}
	Recall that the utility function $\phi(\cdot)$ is concave. Hence, its derivative is non-decreasing in its argument. Furthermore, under the action of any policy, the cumulative reward $R_i(\cdot)$ is non-decreasing for each user $i \in [m].$ Thus, we have $\phi'(R_i(t)) \geq \phi'(R_i(T))$ for all $t \in [T], i\in [m].$ Hence, 
	\begin{eqnarray} \label{bd1}
	 (A) \leq \sum_{t=1}^T \langle \sum_i \phi'(R_i(t))x_i(t), y_i^*\rangle \leq (A'). 
	\end{eqnarray}
	
	\paragraph{\underline{Proving \textcolor{blue}{$(B') \leq (1-\alpha)^{-1}(B)$}}}
	We have 
	\begin{eqnarray} \label{bd2}
		(B)'&=& \sum_{i} \sum_{t=1}^T \phi'(R_i(t)) \langle  x_i(t), y_i(t) \rangle \nonumber\\
		&=& \sum_i \sum_{t=1}^T \phi'(R_i(t)) \big(R_i(t+1)-R_i(t)\big) \nonumber\\
		&\stackrel{(a)}{\leq} & \sum_i \int_{0}^{R_i(T)} \phi'(R) dR \nonumber\\
		&=& \sum_i \phi (R_i(T))\nonumber\\
		&\stackrel{(b)}{=}& (1-\alpha)^{-1}\sum_i \phi'(R_i(T)) R_i(T)\nonumber\\
		&\stackrel{(c)}{=}& (1-\alpha)^{-1} \sum_i \phi'(R_i(T)) \sum_{t=1}^T \langle x_i(t), y_i(t)\rangle  \nonumber\\
		&=& (1-\alpha)^{-1}(B),
	\end{eqnarray}
	where, in (a), we have used the fact that $\phi'(\cdot)$ is non-increasing and $R_i(t+1)-R_i(t) = \langle x_i(t), y_i(t) \leq 1$ (see Figure \ref{ineq-comp} in the Appendix for a geometric visualization), in (b), we have used the explicit form of the $\alpha$-fair utility function to substitute $x \phi'(x) = (1-\alpha)\phi(x)$,
 and in (c), we have used \eqref{hit-eqn}. Combining \eqref{bd1} and \eqref{bd2} and choosing $\beta= (1-\alpha)^{-1},$ we conclude that 
 \begin{eqnarray*}
 	\textrm{Regret}_T((1-\alpha)^{-(1-\alpha)}) \leq (1-\alpha)^\alpha \hat{\textrm{Regret}_T}. 
 \end{eqnarray*}
% Finally, since $\max_{0\leq \alpha <1}(1-\alpha)^{-(1-\alpha)}= e^{1/e},$ the result follows. 
\end{proof}

\subsection{Proof of Lemma \ref{reg-bd-alpha}} \label{reg-bd-alpha-proof}

\begin{proof}
%In this section, we establish that the OGD policy achieves $O(1)$ regret for $\alpha$-fair utility functions.
%for the logarithmic utility function and simultaneously achieves almost linearly many hit rates for all users. 
%the causal online caching policy that solves a causal version of the OLO problem \eqref{olo-red} with $w_i(\tau) \gets \frac{1}{R_i(\tau)}, \forall i$ scores almost linearly many cache hits with respect to time for each user. 
%The causal OCO policy with the above rate can also be seen to be an instance of a \emph{Proportional Fair} policy, that prioritizes the users in inverse proportion to their current hit rates. 

We will be using the following adaptive regret bound for the online gradient descent policy with an appropriate adaptive step sizes sequence. We will see that the norm of the gradients diminishes at a steady rate under the action of the \texttt{OPF} policy. Hence, the data-dependent regret bound plays a central role for a tight regret analysis of the \texttt{OPF} policy.
\begin{framed} 
\begin{theorem}[(Theorem 4.14 of \cite{orabona2019modern})]
	Let $\Delta \subset \mathbb{R}^{d}$ be a convex set with with diameter $D.$ Let us consider a sequence of linear reward functions with gradients $\{\bm{g}_t\}_{t \geq 1}.$ Run the Online Gradient Ascent policy with step sizes $\eta_t= \frac{D}{\sqrt{2\sum_{\tau=1}^t|| \bm{g}_\tau||_2^2}}, 1\leq t \leq T.$ Then the standard regret under the OGA policy can be upper-bounded as follows:
	\begin{eqnarray} \label{data-dep-bd}
	 \textrm{Regret}_T \leq D \sqrt{2\sum_{t=1}^T||\bm{g}_t||_2^2}. 		
	\end{eqnarray}
\end{theorem}  
\end{framed}
Note that, the gradient component $g_i$ corresponding to the $i$\textsuperscript{th} user for the surrogate problem \eqref{surr-regret} on round $t$ is given by the vector $g_i(t)= \phi'(R_i(t)) x_i(t).$ Using the above data-dependent static regret bound \eqref{data-dep-bd}, the regret  achieved by the OGA policy for the surrogate problem for any round $T\geq 1$ can be upper-bounded as follows:
%achieved by the OGD policy with adaptive step sizes \citep[Theorem 4.14]{orabona2019modern},
% \textcolor{red}{(State the theorem and cite the paper)}
%\begin{eqnarray*}
%	\reg = O\bigg(\sqrt{\sum_{t=1}^T ||\bm{g}_t||_2^2}\bigg)
%\end{eqnarray*}
%where $||\bm{g}_t||_2$ denotes the $\ell_2$ norm of the reward function at round $t$. In our case, the above theorem yields 
%
\begin{eqnarray} \label{adaregret}
	\hat{\textrm{Regret}}_T = O\bigg(\sqrt{\sum_{t=1}^T\sum_i \phi'(R_i(t))^2}\bigg) = O\bigg(\sqrt{\sum_{t=1}^T\sum_i \frac{1}{R_i(t)^{2\alpha}}}\bigg),
\end{eqnarray}
where we have used the fact that the demand vectors at each round are bounded.

Clearly, the regret bound \eqref{adaregret} depends on the sequence of the cumulative rewards $\{\bm{R}(t)\}_{t \geq 1}$, which is implicitly controlled by the past actions of the online policy itself. By the definition of regret, 
%By virtue of the sublinear regret property of the OGD policy, 
for any fixed allocation $y^* \in \Delta^N_k,$ we have for any time step $T$: 
\begin{eqnarray} \label{reg-ineq}
	\sum_{t=1}^T \sum_{i} \langle \phi'(R_i(t)) x_i(t), y_i(t) \rangle  \geq \sum_{t=1}^T \sum_i \langle \phi'(R_i(t)) x_i(t), y_i^* \rangle - \hat{\textrm{Regret}}_T, 
\end{eqnarray} 
where $\hat{\textrm{Regret}}_T$ denotes the worst-case cumulative regret of the adaptive OGD policy up to time $T$. Since the cumulative reward of each user is monotonically non-decreasing, we have:
\begin{eqnarray} \label{bd0}
	R_i(T) \geq 1, \forall i, \forall T.
\end{eqnarray}
% This can be ensured by an appropriate initialization such that, at round $0,$ all users request the same file (say file $1$) and the policy caches that file.

This can easily be ensured by an appropriate initialization using Assumptions \ref{assumption1} and \ref{assumption2}.

Substituting this bound to the regret bound in \eqref{adaregret}, we obtain our first (loose) upper-bound for the regret of the fair allocation problem \eqref{surr-regret}:
\begin{eqnarray} \label{reg-bd1}
	\hat{\textrm{Regret}}_T = O(\sqrt{T}).
\end{eqnarray}
Note that this bound might be too loose as the offline benchmark itself could be smaller in magnitude than this regret bound. A closer inspection reveals the root cause for this looseness - the cumulative reward lower bound \eqref{bd0} is too loose for our purpose, as cumulative rewards grow steadily with time, depending on the online policy. We now strengthen the above upper-bound using a novel \emph{bootstrapping} method, that \emph{simultaneously} tightens the lower bound for the cumulative rewards \eqref{bd0} and, consequently, improves the regret upper bound \eqref{adaregret}. 
%\newpage
\iffalse
\hrule
\textbf{Note} The actual regret bound could be better than $O(\sqrt{T}).$ Because if the $R_i(T)'s$ increase linearly, then the magnitude of the gradients increase linearly as well. Upon using the regret bound in terms of the gradients, this yields an $O(1/\sqrt{T})$ regret bound for large $T$. 
\hrule
 \fi
 
 Using the fact that $\langle x_i(t), y_i(t) \rangle = R_i(t+1)-R_i(t),$ and following the same steps as \eqref{bd2}, we have
 \begin{eqnarray}\label{hit-bd1}
 	\sum_{i} \sum_{t=1}^T \phi'(R_i(t)) \langle  x_i(t), y_i(t) \rangle  \leq \sum_i \phi(R_i(T)). 
 \end{eqnarray}
 Furthermore, lower bounding $\phi'(R_i(t))$ by $\phi'(R_i(T)),$ we have
 \begin{eqnarray} \label{hit-bd2}
 	\sum_{t=1}^T \sum_i \langle \phi'(R_i(t)) x_i(t), y_i^* \rangle \geq \sum_i \phi'(R_i(T)) R_i^*(T),
 \end{eqnarray}
 where $R_i^*(T)\equiv \sum_{t=1}^T \langle x_i(t), y_i^*\rangle$ is the cumulative reward accrued by a fixed allocation $y^* \in \Delta$ up to time $T.$  Using Assumptions \ref{assumption1} and \ref{assumption2} and choosing $y_i^*=\mu \mathds{1}_N,$ we have $R_i^*(T) \geq \mu \delta T, \forall i.$ Hence, combining \eqref{hit-bd1}, \eqref{hit-bd2}, and \eqref{reg-ineq}, we have 
 \begin{eqnarray} \label{oco-prop}
 	\sum_i \phi(R_i(T)) \geq \mu \delta T\sum_i \phi'(R_i(T)) - \hat{\textrm{Regret}}_T.
 \end{eqnarray}
 %inequality \eqref{reg-ineq} yields: 
%\begin{eqnarray} \label{oco-prop}
%	\sum_i \log R_i(T) = \sum_i \int_{1}^{R_i(T)} \frac{dR_i(t)}{R_i(t)} 
%	\geq  \sum_i \sum_{t=1}^T \frac{R_i(t+1)-R_i(t)}{R_i(t)} \geq   \sum_i \frac{\textsf{OPT}_i(T)}{R_i(T)} - \reg, 
%\end{eqnarray}
%where we have used the fact that the cumulative hits $R_i(t)$'s are monotone non-decreasing and $\textsf{OPT}_i(T)$ denotes the cumulative hits of the $i$\textsuperscript{th} user obtained by any fixed offline policy $y^*.$ Recall that inequality \eqref{oco-prop} is valid for any fixed caching configuration $y^* \in \Delta^N_k$. Hence, taking uniform average over all fixed cache configurations obtained by randomly selecting each item with uniform probability, we have that \[\textsf{OPT}_i(T) \geq \beta T, \forall i\] for $\beta \equiv \frac{k}{N}>0.$  Hence, the OEGD policy yields the following performance bound:
%\begin{eqnarray} \label{func-ineq1}
%	\frac{1}{T} \sum_i \log R_i(T) \geq \beta \sum_i \frac{1}{R_i(T)} - \frac{\reg}{T}.
%\end{eqnarray}
Since $0<R_i(T) \leq T, \forall i,$ and $\phi(\cdot)$ is monotone non-decreasing, for any user $i \in [m]$, the above inequality yields:
\begin{eqnarray*}
	\frac{m}{(1-\alpha)T^\alpha} \geq \frac{\mu \delta}{R_i^\alpha(T)} - \frac{\hat{\textrm{Regret}}_T}{T}. 
\end{eqnarray*}
\emph{i.e.},
%\begin{eqnarray*}
%	\frac{\reg + m \log T}{T} \geq \frac{k}{N} \frac{1}{R_i(T)}.
%\end{eqnarray*} 
\begin{eqnarray} \label{R-lb}
	R_i^\alpha(T) \geq \mu \delta \bigg(\frac{\hat{\textrm{Regret}}_T}{T} + \frac{m}{1-\alpha}T^{-\alpha}\bigg)^{-1} = \Omega(T^{\min(1/2, \alpha)}), \forall i\in [m],
\end{eqnarray}
where we have used our previous upper bound \eqref{reg-bd1} on the regret. Eqn.\ \eqref{R-lb} is our key equation for carrying out the bootstrapping process as it lower bounds the minimum cumulative reward of users in terms of the worst-case regret. Now we consider two cases:
%, we have, $\reg = O(\sqrt{T}),$ the previous estimate yields:

\begin{figure}
\centering
	\includegraphics[scale=0.4]{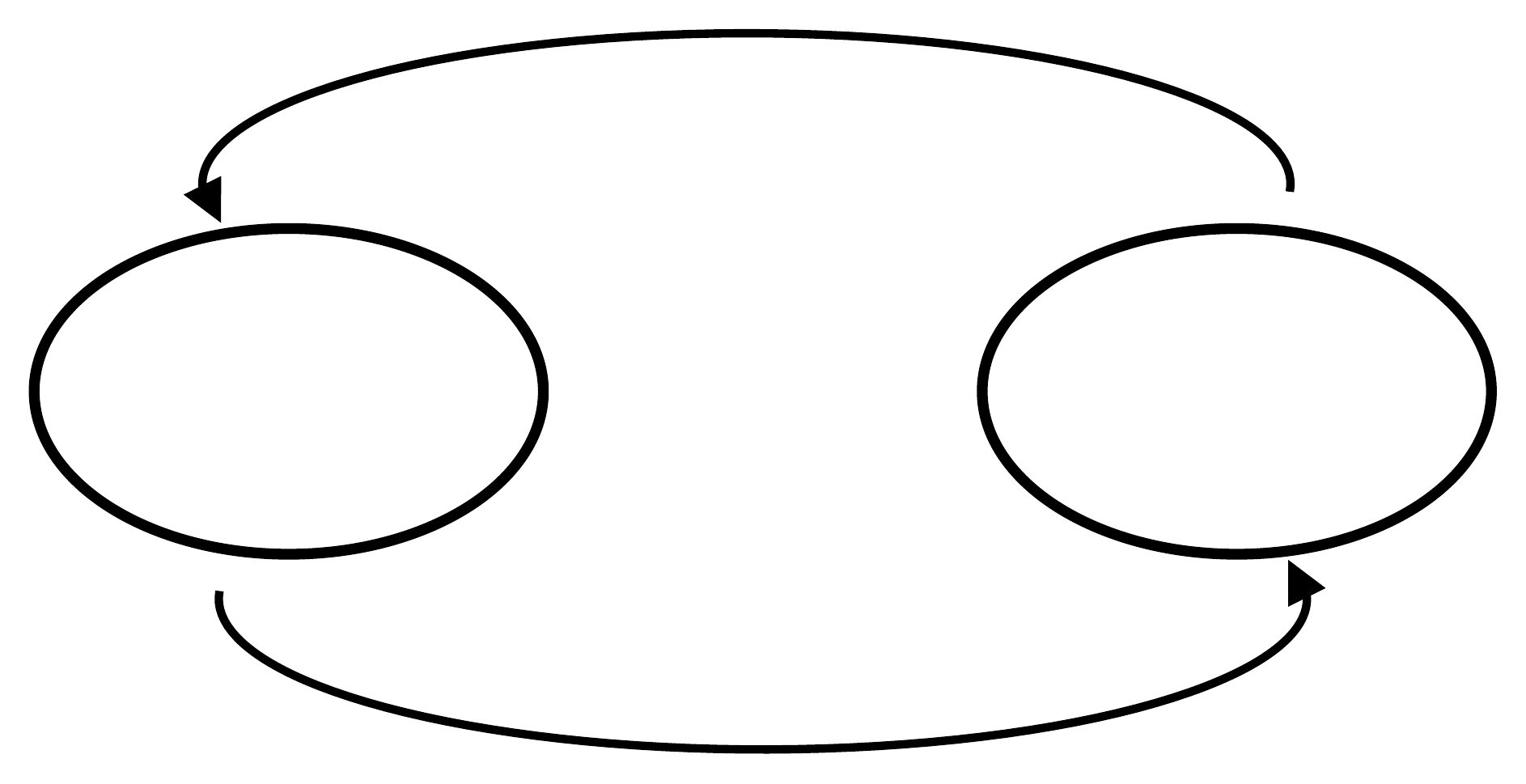}
\put(-66,62){Improve the}
\put(-72,50){lower bounds to}
\put(-58, 38){$R_i(t), \forall i$}
\put(-203,62){Improve the}
\put(-208, 50){upper bound to}
\put(-194, 38){$\hat{\textrm{Regret}_t}$}
\caption{Illustrating the new \emph{Bootstrapping} technique used in the proof of Lemma \ref{reg-bd-alpha}}
\label{bootstrap}
\end{figure}

\textbf{Case-I: $0\leq \alpha \leq \nicefrac{1}{2}$}

In this case, we immediately have $R_i(T)=\Omega(T), \forall i,T.$ Substituting this bound on \eqref{adaregret}, we have 
\begin{eqnarray*}
	\hat{\textrm{Regret}}_T  = \begin{cases}
		O(\sqrt{\log T})~~ \textrm{if} ~~\alpha=1/2\\
		O(T^{1/2-\alpha}) ~~ \textrm{if} ~~0<\alpha <1/2.
	\end{cases}
\end{eqnarray*}
\textbf{Case-II: $\nicefrac{1}{2} < \alpha <1$}

In this case, using the bound \eqref{R-lb}, we have $R_i(T)=\Omega(T^{1/2\alpha}), \forall i,T.$ Using the regret bound \eqref{adaregret}, the regret can be bounded as
\begin{eqnarray*}
	\hat{\textrm{Regret}}_T = O(\sqrt{\sum_{t=1}^T \frac{1}{t}})= O(\sqrt{\log T}).
\end{eqnarray*}
Substituting this bound again in \eqref{R-lb}, we have $R_i(T) \geq \Omega(T).$ This, in turn, yields the following regret bound:
\begin{eqnarray*}
	\hat{\textrm{Regret}}_T = O\big(\sqrt{\sum_{t=1}^T \frac{1}{t^{2\alpha}}}\big)=O(1).
\end{eqnarray*}

%\begin{eqnarray} \label{bd1}
%	R_i(T) = \Omega(\sqrt{T}).
%\end{eqnarray}
%%This implies that the difference between two successive gradients is 
%%\begin{eqnarray*}
%%	 ||\sum_i \frac{x_i(t)}{R_i(t)} - \sum_i \frac{x_i(t-1)}{R_i(t-1)}||_2 \stackrel{\textrm{(Triangle ineq.)}}{\leq} \sum_i O(\frac{1}{\sqrt{t}})= O(\frac{1}{t^{1/2}}).
%%\end{eqnarray*}
%Upon combining \eqref{adaregret} with the new lower bound \eqref{bd1} for the cumulative hits, and using Theorem \ref{adaptive-regret}, we can now boost the previous regret bound \eqref{reg-bd1} to arrive at the following tighter bound: 
%\begin{eqnarray} \label{reg-bd2}
%	\reg = O\bigg(\sqrt{\sum_{t=1}^T} \frac{1}{t}\bigg)= O(\sqrt{\log T}).
%\end{eqnarray}
%Substituting the new regret bound in \eqref{func-ineq1}, we arrive at the following lower bound for cumulative hits:
%\begin{eqnarray*}
%	R_i(T) = \Omega\big(\frac{T}{\log T}\big), \forall i. 
%\end{eqnarray*} 
%Using the regret bound \eqref{adaregret} one more time, we have the following final upper bound on regret
%\begin{eqnarray*}
%	\textrm{Reg}(T) = O(\sqrt{\sum_{t=1}^T \frac{(\log t)^2}{t^2}})=O(1). 
%\end{eqnarray*}
\end{proof}

\subsection{Simplified Pseudocode for the Online Proportional Fair Policy for the Caching Problem} \label{simplified}

  \begin{algorithm}[H]
\caption{Online Proportional Fair Caching (\texttt{OPFC})}
\label{OPFC}
\begin{algorithmic}[1]
\State \algorithmicrequire{ Fairness parameter $0\leq \alpha<1.$} Online file request sequence $\{x_i(t)\}_{t=1}^T$ from user $i, i \in [m],$ Euclidean projection oracle $\Pi_{\Delta^N_k}(\cdot)$ onto the feasible convex set $\Delta^N_k.$
\State \algorithmicensure{ Sequence of file inclusion probabilities $\{y_t\}_{t=1}^T$ and samples from this distribution}
\State $R_i \gets 1, \forall i\in [m], \bm{y}\gets \frac{k}{N}\mathds{1}_{N}, S\gets 0.$ \algorithmiccomment{\textcolor{black}{\emph{Initialization}}}
\ForEach {round $t=2:T$:}
\State $g \gets \sum_{i=1}^m \frac{x_i(t-1)}{R_i^\alpha}$ \algorithmiccomment{\emph{Computing the gradient}}
\State $S \gets S+ ||g||_2^2,$   \algorithmiccomment{\emph{Accumulating the norm of the gradients}}
\State $y \gets \Pi_{\Delta^N_k}\big(y+ \frac{k}{\sqrt{S}} g\big)$ \algorithmiccomment{\textcolor{black}{\emph{Updating the inclusion probabilities using OGA}}}
\State The users reveal their latest file requests $\{x_i(t), i \in [m]\}.$
\State $R_i \gets R_i + \langle x_i(t), y \rangle, ~\forall i\in [m].$ \algorithmiccomment{\textcolor{black}{\emph{Updating cumulative hits}}}
%\State Cache file $S_t \gets \textsc{Madow-Sample}(y)$
%$k$ files according to the inclusion probability vector $y$ using Madow's sampling scheme described in Algorithm \ref{madow}
%given in Algorithm \ref{uneq} 
\algorithmiccomment{\textcolor{black}{\emph{Sample a subset of $k$ files by invoking Algorithm \ref{Madow}}}}
%according to the \textsc{FTRL}($\eta$) update 
%\eqref{ftrl-opt}. 
%\STATE Feed the vector $\bm{g}_t$ as the reward vector for round $t$ to the \textsc{FTRL} policy 
\EndForEach
\end{algorithmic}
\end{algorithm}

\subsection{Efficiently sampling an integral allocation from a mixed allocation in the convex hull} \label{sampling-tech}
\paragraph{1. Online shared caching:}
For the online shared caching problem, the admissible set $\Delta$ is given by Eq.\ \eqref{caching-feasible}. Clearly, incidence vectors of all $k$-sets, containing $k$ 1's and $N-k$ zeros, belong to $\Delta.$ Furthermore, given any vector in $\bm{y} \in \Delta,$ a randomized integral allocation can be sampled in linear time using Madow's sampling scheme, described in Algorithm \ref{Madow}, which yields the vector $\bm{y}$ in expectation. 

\begin{algorithm} [h]
\caption{\textsc{Madow-Sample} ($p$)}
\label{Madow}
\begin{algorithmic}[1]
\State \algorithmicrequire{ A universe $[N]$ of size $N$, the cardinality of the sampled set $k$, and a  marginal inclusion probability vector $p \in \Delta^N_k.$}
\State \algorithmicensure{ A random $k$-set $S$ with $|S|=k$ such that, $\mathbb{P}(i \in S)=p_{i}, \forall i\in [N]$} 
\State Define $\Pi_0=0$, and $\Pi_i= \Pi_{i-1}+p_{i}, \forall 1\leq i \leq N.$
\State Sample a uniformly distributed random variable $U$ from the interval $[0,1].$
\State $S \gets \emptyset$
\ForEach {$i\gets 0$ to $k-1$}
\State Select the element $j$ if $\Pi_{j-1} \leq U + i < \Pi_j.$ 
%\State $S \gets S \cup \{j\}.$
\EndForEach
\State \Return $S$
\end{algorithmic}
\end{algorithm}

\paragraph{2. Online job scheduling:} 
In the online job scheduling problem the admissible action set $\Delta$ is given by the $N$-simplex. Given any point on the $N$-simplex, it is trivial to randomly sample a coordinate using a single uniform random variable.

\paragraph{3. Online matching:}

In the online matching problem, the admissible action set $\Delta$ is given by the Birkhoff polytope \eqref{bvn-polytope}. Given any point $\bm{y}$ in the Birkhoff polytope, it can be efficiently decomposed as a convex combination of a small number of matchings using the Birkhoff-von-Neumann (BvN) decomposition algorithm \citep{valls2021birkhoff}. Using the coefficients of the decomposition, a matching can be randomly sampled that exactly matches the point $\bm{y}$ in expectation.    

\subsection{Non-Convexity of the optimal offline benchmark for the online job scheduling problem} \label{non-cvx}

Consider the job scheduling problem in the reward maximization setting as described in Example \ref{sched-example}. Clearly, the $\alpha$-fairness metric accumulated by the online policy is concave in the regime $0\leq \alpha \leq 1.$ However, the following proposition shows that the offline static benchmark for this problem is \emph{non-convex} for the $\alpha$-fair utility function in the regime $0<\alpha <1.$
\begin{framed}
\begin{proposition}
	The offline optimal $\alpha$-fair reward function for the job scheduling problem is non-convex for $0<\alpha<1.$
\end{proposition}
\end{framed}
% and the $(1-\alpha)$-norm reward function (in the regime $1/2<\alpha <1$). 
%\paragraph{Case-I: $\alpha$-fair Utility function:}
\begin{proof}
Let the probability distribution $\bm{y}^*$ be an optimal static offline allocation vector for the reward sequence $\{\bm{x}_t\}_{t \geq 1}.$ Also, let $R_T(i)\equiv \sum_t x_i(t)$ be the cumulative reward observed by the $i$\textsuperscript{th} machine, $1\leq i \leq m$. Then the optimal allocation $\bm{y}^*$ maximizes the following objective:
\begin{eqnarray} \label{job-obj}
	(1-\alpha)^{-1} \sum_i (y_i R_T(i))^{1-\alpha}. 
\end{eqnarray}
s.t. the constraints $\sum_i y_i=1, \bm{y} \geq \bm{0}.$ Since $0<\alpha<1,$ the objective function is strongly concave and the optimal solution $\bm{y}^*$ is unique. Using the standard H\"older's inequality with the conjugate norms $p=\nicefrac{1}{1-\alpha}, q= \nicefrac{1}{\alpha},$ we can upper bound the objective \eqref{job-obj} as 
\begin{eqnarray} \label{max-obj1}
	(1-\alpha)^{-1}\sum_{i=1}^m (y_i R_T(i))^{1-\alpha}  &\leq& (1-\alpha)^{-1}(\sum_i y_i)^{1-\alpha} (\sum_i R_T(i)^{\frac{1-\alpha}{\alpha}})^{\alpha}\nonumber \\
	&=& (1-\alpha)^{-1} (\sum_i R_T(i)^{\frac{1-\alpha}{\alpha}})^{\alpha},
\end{eqnarray}
where we have used the constraint $\sum_i y_i=1$ in the last equality. 
The upper-bound \label{max-obj1} is achieved by the following distribution 
\begin{eqnarray} \label{max-dist}
	y^*_i =  \frac{R_i^{\frac{1-\alpha}{\alpha}}}{\sum_i R_i^{\frac{1-\alpha}{\alpha}}}, ~\forall i \in [m].
\end{eqnarray}
Hence, the optimal offline reward \eqref{job-obj} is given by 
\begin{eqnarray} \label{offline-reward}
\mathcal{R}^* = (1-\alpha)^{-1} (\sum_i R_T(i)^{\frac{1-\alpha}{\alpha}})^{\alpha}.	
\end{eqnarray}
 
To show that $\mathcal{R}^*$ is non-convex in the cumulative reward vector $\bm{R}_T$ in the regime $0<\alpha <1,$ consider the case of two users. Letting $x=R_T(1), y=R_T(2)$ and ignoring the positive pre-factor, the function under consideration is given as follows:
\begin{eqnarray} \label{opt-fun}
	f(x,y)= (x^{\frac{1-\alpha}{\alpha}}+y^{\frac{1-\alpha}{\alpha}})^{\alpha}.
\end{eqnarray}
Recall that a function is convex if and only if the determinants of all leading principal minors of its Hessian matrix are non-negative. A straightforward computation yields the following expression for the determinant of the Hessian of the function $f(x,y):$
\begin{eqnarray*}
	\textrm{det}(\nabla^2 f(x,y)) = (2 \alpha - 1)\frac{(\alpha - 1)^2  x^{1/\alpha - 1} y^{1/\alpha - 1} (x^{1/\alpha - 1}+ y^{1/\alpha - 1})^{2 \alpha}}{(y x^{1/\alpha} + x y^{1/\alpha})^2}.
\end{eqnarray*}
Clearly, the determinant becomes strictly negative in the regime $0<\alpha <\frac{1}{2}.$ This shows that the function \eqref{opt-fun} is non-convex for $0<\alpha <\frac{1}{2}$. On the other hand, we have 
\begin{eqnarray*}
	\frac{\partial^2 f(x,y)}{\partial x^2} = (\alpha - 1)\frac{y x^{1/\alpha - 2} (x^{1/\alpha - 1} + y^{1/\alpha - 1})^\alpha (\alpha^2 y x^{1/\alpha} + 2 \alpha x y^{1/\alpha} - x y^{1/\alpha})}{\alpha (y x^{1/\alpha} + x y^{1/\alpha})^2}. 
\end{eqnarray*}
In particular, at the point $(1,1)$ the above second partial derivative evaluates to be  
\begin{eqnarray*}
		\frac{\partial^2 f(x,y)}{\partial x^2}|_{(x,y)=(1,1)} = -\frac{2^{\alpha-2}(1-\alpha)}{\alpha}(\alpha^2+2\alpha-1),
\end{eqnarray*}
which is strictly negative for $\sqrt{2}-1< \alpha <1.$ Taking the above two results together, we conclude that the offline optimal reward function \eqref{opt-fun} is non-convex for  $0<\alpha <1.$ 
\end{proof}
%\edit{I computed the above expression using Wolfram alpha. Ativ, can you please verify quickly the above by computing these by hand?}

\subsection{Proof of Theorem \ref{lb-thm} (lower bounding the approximation factor)} \label{lb-proof}

%In this section, we will derive a lower bound on the ratio between the offline optimal hit rates and the hit rate of any online algorithm. This gives a lower bound on the value of $c$ in the definition of $c$-regret. 
Consider the following instance of the online shared caching problem with $m=2$ users and a cache of unit capacity. Assume that size (\(N \)) of the library is sufficiently large. Let the total length of the request sequence be \(T \) rounds and let \(\eta \in [0,1]\) be some constant fraction to be fixed later. Consider two different file request sequence:

\begin{itemize}
    \item \textbf{Instance 1:} For the first \(\eta T \) rounds, user 1 always requests file 1 and user 2 always requests file 2. For the next \((1-\eta)T \) rounds, user 1 always requests file 2 and user 2 requests a file chosen uniformly at random from the library.
    \item \textbf{Instance 2:} For the first \(\eta T \) rounds, as in Instance $1$, user 1 always requests file 1 and user 2 always requests file 2. However, for the next \((1-\eta)T \) rounds, user 1 requests a random file chosen uniformly at random from the library and user 2 always requests file 1.
\end{itemize}

%Our goal is to lower bound the achievable approximation ratio for the above fair caching problem.
We now lower bound the approximation factor achievable by any online policy for the above two request sequences.

\textbf{Optimal static offline policy:} It is easy to see that the optimal offline strategy is to cache file $2$ for all $T$ rounds. Hence, the total $\alpha$-fairness objective accrued by the static offline policy is \[ \textsf{Offline $\alpha$-fairness}=(1-\alpha)^{-1}T^{1-\alpha}\big[\eta^{1-\alpha} + (1-\eta)^{1-\alpha}\big].\] Clearly, the same fairness objective is achieved for the second instance by a static policy that always caches file 1.

\textbf{Online Policy:} Suppose that during the first \(\eta T \) rounds, the online policy caches file 1 for \(\gamma\) fraction of the time and file 2 for \(1-\gamma\) fraction of the time. Since the policy is online, the fraction $\gamma$ remains the same for both instances. Clearly, for the next \((1-\eta)T \) time slots, an optimal online policy caches file 2 for instance 1 and file 1 for instance 2. Taking worst among the two instances, any online policy achieves the following fairness objective: 
\begin{eqnarray*}
	&&\textsf{Online $\alpha$-fairness} = (1-\alpha)^{-1}T^{1-\alpha} \min \big(g(\gamma), g(1-\gamma)\big),\\
	 %&&\min\bigg(\big(\gamma\eta T + (1-\eta)T\big)^{1-\alpha} +\big((1-\gamma)\eta T\big)^{1-\alpha}, \big((1-\gamma)\eta T + (1-\eta)T\big)^{1-\alpha} + \big(\gamma\eta T\big)^{1-\alpha} \bigg). 
\end{eqnarray*}
where $g(\gamma) \equiv \big(\gamma\eta + (1-\eta)\big)^{1-\alpha} +\big((1-\gamma)\eta \big)^{1-\alpha}.$ From simple calculus, it follows that the function $g(\gamma)$ is non-increasing for $0\leq \eta \leq \nicefrac{1}{2}, 0\leq \alpha<1.$ This implies that the online $\alpha$-fairness is maximized when \(\gamma^* = \frac{1}{2}\). Hence, the $\alpha$-fairness metric achieved by any online policy for the worst of the above two instances is upper bounded by: 
\begin{eqnarray*}
	\textsf{Online $\alpha$-fairness} \leq {(1-\eta/2) }^{1-\alpha} + {({\eta }/{2})}^{1-\alpha}.
\end{eqnarray*}

%\[\text{Online}=\sqrt{\frac{\eta T }{2} + (1-\eta)T } + \sqrt{\frac{\eta T}{2}}.\]
%
%Hence, the approximation ratio is lower bounded by taking \(\eta = 0.2\):
%
%\begin{equation}
%    \texttt{ratio}\geq\frac{\text{offline}}{\text{online}}= \frac{\sqrt{\eta } + \sqrt{(1-\eta)}}{\sqrt{1-\eta/2 } + \sqrt{{\eta }/{2}}} = \frac{3}{2\sqrt{2}} \approx 1.06.
%\end{equation}

%Following the same steps as above, we can show that for an arbitrary $\alpha$, the ratio is lower bounded by 
Hence, the achievable approximation ratio $c_\alpha$ for any online policy with a sublinear $c_\alpha$-regret is lower bounded as follows:
\begin{equation}
    \textsf{Approx.\ ratio}\geq\frac{\textsf{Offline $\alpha$-fairness}}{\textsf{Online $\alpha$-fairness}}= \frac{{\eta }^{1-\alpha} + {(1-\eta)}^{1-\alpha}}{{(1-\eta/2) }^{1-\alpha} + {({\eta }/{2})}^{1-\alpha}}.
\end{equation}
Taking the maximum of the RHS with respect to the parameter $\eta$ yields the desired result.

\clearpage
\bibliography{OCO}

\begin{thebibliography}{29}
\providecommand{\natexlab}[1]{#1}
\providecommand{\url}[1]{\texttt{#1}}
\expandafter\ifx\csname urlstyle\endcsname\relax
  \providecommand{\doi}[1]{doi: #1}\else
  \providecommand{\doi}{doi: \begingroup \urlstyle{rm}\Url}\fi

\bibitem[Even-Dar et~al.(2009)Even-Dar, Kleinberg, Mannor, and
  Mansour]{even2009online}
Eyal Even-Dar, Robert Kleinberg, Shie Mannor, and Yishay Mansour.
\newblock Online learning for global cost functions.
\newblock In \emph{COLT}, 2009.

\bibitem[Dwork et~al.(2012)Dwork, Hardt, Pitassi, Reingold, and
  Zemel]{dwork-fairness}
Cynthia Dwork, Moritz Hardt, Toniann Pitassi, Omer Reingold, and Richard Zemel.
\newblock Fairness through awareness.
\newblock In \emph{Proceedings of the 3rd Innovations in Theoretical Computer
  Science Conference}, ITCS '12, page 214–226, New York, NY, USA, 2012.
  Association for Computing Machinery.
\newblock ISBN 9781450311151.
\newblock \doi{10.1145/2090236.2090255}.
\newblock URL \url{https://doi.org/10.1145/2090236.2090255}.

\bibitem[Hao(2019)]{hao2019facebook}
Karen Hao.
\newblock Facebook’s ad-serving algorithm discriminates by gender and race.
\newblock \emph{MIT Technology Review}, 2019.

\bibitem[Lan et~al.(2010)Lan, Kao, Chiang, and Sabharwal]{lan2010axiomatic}
Tian Lan, David Kao, Mung Chiang, and Ashutosh Sabharwal.
\newblock \emph{An axiomatic theory of fairness in network resource
  allocation}.
\newblock IEEE, 2010.

\bibitem[Kelly(1997)]{kelly1997charging}
Frank Kelly.
\newblock Charging and rate control for elastic traffic.
\newblock \emph{European transactions on Telecommunications}, 8\penalty0
  (1):\penalty0 33--37, 1997.

\bibitem[Mo and Walrand(2000)]{mo2000fair}
Jeonghoon Mo and Jean Walrand.
\newblock Fair end-to-end window-based congestion control.
\newblock \emph{IEEE/ACM Transactions on networking}, 8\penalty0 (5):\penalty0
  556--567, 2000.

\bibitem[Radunovic and Le~Boudec(2007)]{radunovic2007unified}
Bozidar Radunovic and Jean-Yves Le~Boudec.
\newblock A unified framework for max-min and min-max fairness with
  applications.
\newblock \emph{IEEE/ACM Transactions on networking}, 15\penalty0 (5):\penalty0
  1073--1083, 2007.

\bibitem[Nace and Pi{\'o}ro(2008)]{nace2008max}
Dritan Nace and Michal Pi{\'o}ro.
\newblock Max-min fairness and its applications to routing and load-balancing
  in communication networks: a tutorial.
\newblock \emph{IEEE Communications Surveys \& Tutorials}, 10\penalty0
  (4):\penalty0 5--17, 2008.

\bibitem[Jain et~al.(1984)Jain, Chiu, Hawe, et~al.]{jain1984quantitative}
Rajendra~K Jain, Dah-Ming~W Chiu, William~R Hawe, et~al.
\newblock A quantitative measure of fairness and discrimination.
\newblock \emph{Eastern Research Laboratory, Digital Equipment Corporation,
  Hudson, MA}, 21, 1984.

\bibitem[Altman et~al.(2010)Altman, Avrachenkov, and Garnaev]{ALTMAN2010338}
E.~Altman, K.~Avrachenkov, and A.~Garnaev.
\newblock Fair resource allocation in wireless networks in the presence of a
  jammer.
\newblock \emph{Performance Evaluation}, 67\penalty0 (4):\penalty0 338--349,
  2010.
\newblock ISSN 0166-5316.
\newblock \doi{https://doi.org/10.1016/j.peva.2009.08.002}.
\newblock URL
  \url{https://www.sciencedirect.com/science/article/pii/S0166531609001035}.
\newblock Performance Evaluation Methodologies and Tools: Selected Papers from
  VALUETOOLS 2008.

\bibitem[Hazan(2019)]{hazan2019introduction}
Elad Hazan.
\newblock Introduction to online convex optimization.
\newblock \emph{arXiv preprint arXiv:1909.05207}, 2019.

\bibitem[Si~Salem et~al.(2022)Si~Salem, Iosifidis, and Neglia]{si2022enabling}
Tareq Si~Salem, Georgios Iosifidis, and Giovanni Neglia.
\newblock Enabling long-term fairness in dynamic resource allocation.
\newblock \emph{Proceedings of the ACM on Measurement and Analysis of Computing
  Systems}, 6\penalty0 (3):\penalty0 1--36, 2022.

\bibitem[Rakhlin et~al.(2011)Rakhlin, Sridharan, and Tewari]{rakhlin2011online}
Alexander Rakhlin, Karthik Sridharan, and Ambuj Tewari.
\newblock Online learning: Beyond regret.
\newblock In \emph{Proceedings of the 24th Annual Conference on Learning
  Theory}, pages 559--594. JMLR Workshop and Conference Proceedings, 2011.

\bibitem[Wang et~al.(2022)Wang, Ye, Lin, and Lui]{online-fair}
Zhiyuan Wang, Jiancheng Ye, Dong Lin, and John C.~S. Lui.
\newblock Achieving efficiency via fairness in online resource allocation.
\newblock In \emph{Proceedings of the Twenty-Third International Symposium on
  Theory, Algorithmic Foundations, and Protocol Design for Mobile Networks and
  Mobile Computing}, MobiHoc '22, page 101–110, New York, NY, USA, 2022.
  Association for Computing Machinery.
\newblock ISBN 9781450391658.
\newblock \doi{10.1145/3492866.3549724}.
\newblock URL \url{https://doi.org/10.1145/3492866.3549724}.

\bibitem[Blum and Burch(1997)]{blum1997line}
Avrim Blum and Carl Burch.
\newblock On-line learning and the metrical task system problem.
\newblock In \emph{Proceedings of the tenth annual conference on Computational
  learning theory}, pages 45--53, 1997.

\bibitem[Blum et~al.(1999)Blum, Burch, and Kalai]{fine-paging}
A.~Blum, C.~Burch, and A.~Kalai.
\newblock Finely-competitive paging.
\newblock In \emph{40th Annual Symposium on Foundations of Computer Science
  (Cat. No.99CB37039)}, pages 450--457, 1999.
\newblock \doi{10.1109/SFFCS.1999.814617}.

\bibitem[Patil et~al.(2021)Patil, Ghalme, Nair, and Narahari]{narahari-fair}
Vishakha Patil, Ganesh Ghalme, Vineet Nair, and Y.~Narahari.
\newblock Achieving fairness in the stochastic multi-armed bandit problem.
\newblock \emph{J. Mach. Learn. Res.}, 22\penalty0 (1), jan 2021.
\newblock ISSN 1532-4435.

\bibitem[Joseph et~al.(2016)Joseph, Kearns, Morgenstern, and
  Roth]{joseph2016fairness}
Matthew Joseph, Michael Kearns, Jamie~H Morgenstern, and Aaron Roth.
\newblock Fairness in learning: Classic and contextual bandits.
\newblock \emph{Advances in neural information processing systems}, 29, 2016.

\bibitem[Li et~al.(2019)Li, Liu, and Ji]{li2019combinatorial}
Fengjiao Li, Jia Liu, and Bo~Ji.
\newblock Combinatorial sleeping bandits with fairness constraints.
\newblock \emph{IEEE Transactions on Network Science and Engineering},
  7\penalty0 (3):\penalty0 1799--1813, 2019.

\bibitem[Bertsimas et~al.(2012)Bertsimas, Farias, and
  Trichakis]{bertsimas2012efficiency}
Dimitris Bertsimas, Vivek~F Farias, and Nikolaos Trichakis.
\newblock On the efficiency-fairness trade-off.
\newblock \emph{Management Science}, 58\penalty0 (12):\penalty0 2234--2250,
  2012.

\bibitem[Bhattacharjee et~al.(2020)Bhattacharjee, Banerjee, and
  Sinha]{SIGMETRICS20}
Rajarshi Bhattacharjee, Subhankar Banerjee, and Abhishek Sinha.
\newblock Fundamental limits on the regret of online network-caching.
\newblock \emph{Proc. ACM Meas. Anal. Comput. Syst.}, 4\penalty0 (2), June
  2020.
\newblock \doi{10.1145/3392143}.
\newblock URL \url{https://doi.org/10.1145/3392143}.

\bibitem[Mhaisen et~al.(2022)Mhaisen, Sinha, Paschos, and
  Iosifidis]{mhaisen2022optimistic}
Naram Mhaisen, Abhishek Sinha, Georgios Paschos, and Georgios Iosifidis.
\newblock Optimistic no-regret algorithms for discrete caching.
\newblock \emph{arXiv preprint arXiv:2208.06414}, 2022.

\bibitem[Joshi and Sinha(2022)]{ITW22}
Ativ Joshi and Abhishek Sinha.
\newblock Universal caching.
\newblock In \emph{2022 IEEE Information Theory Workshop (ITW)}, pages
  684--689, 2022.
\newblock \doi{10.1109/ITW54588.2022.9965906}.

\bibitem[Ziegler(2012)]{ziegler2012lectures}
G{\"u}nter~M Ziegler.
\newblock \emph{Lectures on polytopes}, volume 152.
\newblock Springer Science \& Business Media, 2012.

\bibitem[Tu et~al.(2014)Tu, Ribeiro, Jensen, Towsley, Liu, Jiang, and
  Wang]{tu2014online}
Kun Tu, Bruno Ribeiro, David Jensen, Don Towsley, Benyuan Liu, Hua Jiang, and
  Xiaodong Wang.
\newblock Online dating recommendations: matching markets and learning
  preferences.
\newblock In \emph{Proceedings of the 23rd international conference on world
  wide web}, pages 787--792, 2014.

\bibitem[Orabona(2019)]{orabona2019modern}
Francesco Orabona.
\newblock A modern introduction to online learning.
\newblock \emph{arXiv preprint arXiv:1912.13213}, 2019.

\bibitem[Cesa-Bianchi and Lugosi(2006)]{cesa2006prediction}
Nicolo Cesa-Bianchi and G{\'a}bor Lugosi.
\newblock \emph{Prediction, learning, and games}.
\newblock Cambridge university press, 2006.

\bibitem[Bhaskara et~al.(2020)Bhaskara, Cutkosky, Kumar, and
  Purohit]{bhaskara2020online}
Aditya Bhaskara, Ashok Cutkosky, Ravi Kumar, and Manish Purohit.
\newblock Online learning with imperfect hints.
\newblock In \emph{International Conference on Machine Learning}, pages
  822--831. PMLR, 2020.

\bibitem[Valls et~al.(2021)Valls, Iosifidis, and Tassiulas]{valls2021birkhoff}
V{\'\i}ctor Valls, George Iosifidis, and Leandros Tassiulas.
\newblock Birkhoff’s decomposition revisited: Sparse scheduling for
  high-speed circuit switches.
\newblock \emph{IEEE/ACM Transactions on Networking}, 29\penalty0 (6):\penalty0
  2399--2412, 2021.

\end{thebibliography}
\bibliographystyle{unsrtnat}

\end{document}